\documentclass{article}
\usepackage{}
\usepackage{amsmath, algorithm, graphicx, subcaption, siunitx, float, hyperref, cleveref, amssymb}
\usepackage[noend]{algpseudocode}
\usepackage[utf8]{inputenc}
\usepackage{booktabs}
\usepackage[blocks]{authblk}

\makeatletter
\def\BState{\State\hskip-\ALG@thistlm}
\makeatother

\hypersetup{
    colorlinks=true,
    linkcolor=black,
    citecolor=black,
    urlcolor=blue,
}

\usepackage{natbib}
\setcitestyle{authoryear,open={(},close={)}}

\newcommand{\changeurlcolor}[1]{\hypersetup{urlcolor=#1}} 

\nocite{*}

\title{{\Large Training Verifiers to Solve Math Word Problems}}

\newcommand\CoAuthorMark{\footnotemark[\arabic{footnote}]}
\author{
    \textbf{\scriptsize Karl Cobbe\footnote{Equal contribution. Correspondence to: Karl Cobbe \textless karl@openai.com\textgreater, Vineet Kosaraju \textless vineet@openai.com\textgreater}\hspace{7mm}}
    \textbf{\scriptsize Vineet Kosaraju\protect\CoAuthorMark \hspace{7mm}}
    \textbf{\scriptsize Mohammad Bavarian\hspace{7mm}}
    \textbf{\scriptsize Mark Chen\hspace{5mm}}
    \textbf{\scriptsize Heewoo Jun\hspace{8mm}}
    \textbf{\scriptsize Łukasz Kaiser\hspace{8mm}}
    \textbf{\scriptsize Matthias Plappert\hspace{8mm}}
    \textbf{\scriptsize Jerry Tworek\hspace{8mm}}
    \textbf{\scriptsize Jacob Hilton\hspace{5mm}}
    \textbf{\scriptsize Reiichiro Nakano\hspace{5mm}}
    \textbf{\scriptsize Christopher Hesse\hspace{5mm}}
    \textbf{\scriptsize John Schulman\hspace{5mm}}
}

\affil{\small OpenAI}
\date{}

\begin{document}

\maketitle

\vspace{-.5cm}
\begin{abstract}

State-of-the-art language models can match human performance on many tasks, but they still struggle to robustly perform multi-step mathematical reasoning. To diagnose the failures of current models and support research, we introduce GSM8K, a dataset of 8.5K high quality linguistically diverse grade school math word problems. We find that even the largest transformer models fail to achieve high test performance, despite the conceptual simplicity of this problem distribution. To increase performance, we propose training verifiers to judge the correctness of model completions. At test time, we generate many candidate solutions and select the one ranked highest by the verifier. We demonstrate that verification significantly improves performance on GSM8K, and we provide strong empirical evidence that verification scales more effectively with increased data than a finetuning baseline.

\end{abstract}

\section{Introduction}

In recent years, large language models have demonstrated impressive skills across many diverse tasks \citep{wang2019superglue, brown2020language}. \cite{kaplan2020scaling} describe the consistent benefits of increasing model size, characterizing scaling trends that hold across many orders of magnitude. However, even the largest models falter when required to perform multi-step mathematical reasoning \citep{hendrycks2021measuring}. Model samples frequently contain catastrophic mistakes, even after the model has been appropriately finetuned. Mathematical reasoning thus reveals a critical weakness in modern language models.

One significant challenge in mathematical reasoning is the high sensitivity to individual mistakes \citep{shen2021generate}. When generating a solution, autoregressive models have no mechanism to correct their own errors. Solutions that veer off-course quickly become unrecoverable. If we rely purely on generative methods and extrapolate from current trends, we will require an exorbitant parameter count to achieve even moderate performance on distributions as challenging as the MATH dataset \citep{hendrycks2021measuring}. This evidence strongly motivates the search for methods with more favorable scaling laws.

We propose training verifiers to evaluate the correctness of model generated solutions, similar to concurrent work by \cite{shen2021generate}. At test time, we sample a fixed number of candidate solutions and select the solution ranked highest by the verifier. Verifiers benefit both from their inherent optionality and from verification being a simpler task than generation in general.

To facilitate research, we are releasing GSM8K, a dataset of 8.5K high quality problems at the grade school math level. We designed this dataset to have high linguistic diversity while relying on relatively simple grade school math concepts. State-of-the-art language models struggle to achieve high performance on this dataset, primarily due to the high diversity among problems. At the same time, GSM8K solutions depend only on elementary concepts, so achieving high test performance is a tractable goal.

\begin{figure*}
\centering
\includegraphics[width=\textwidth]{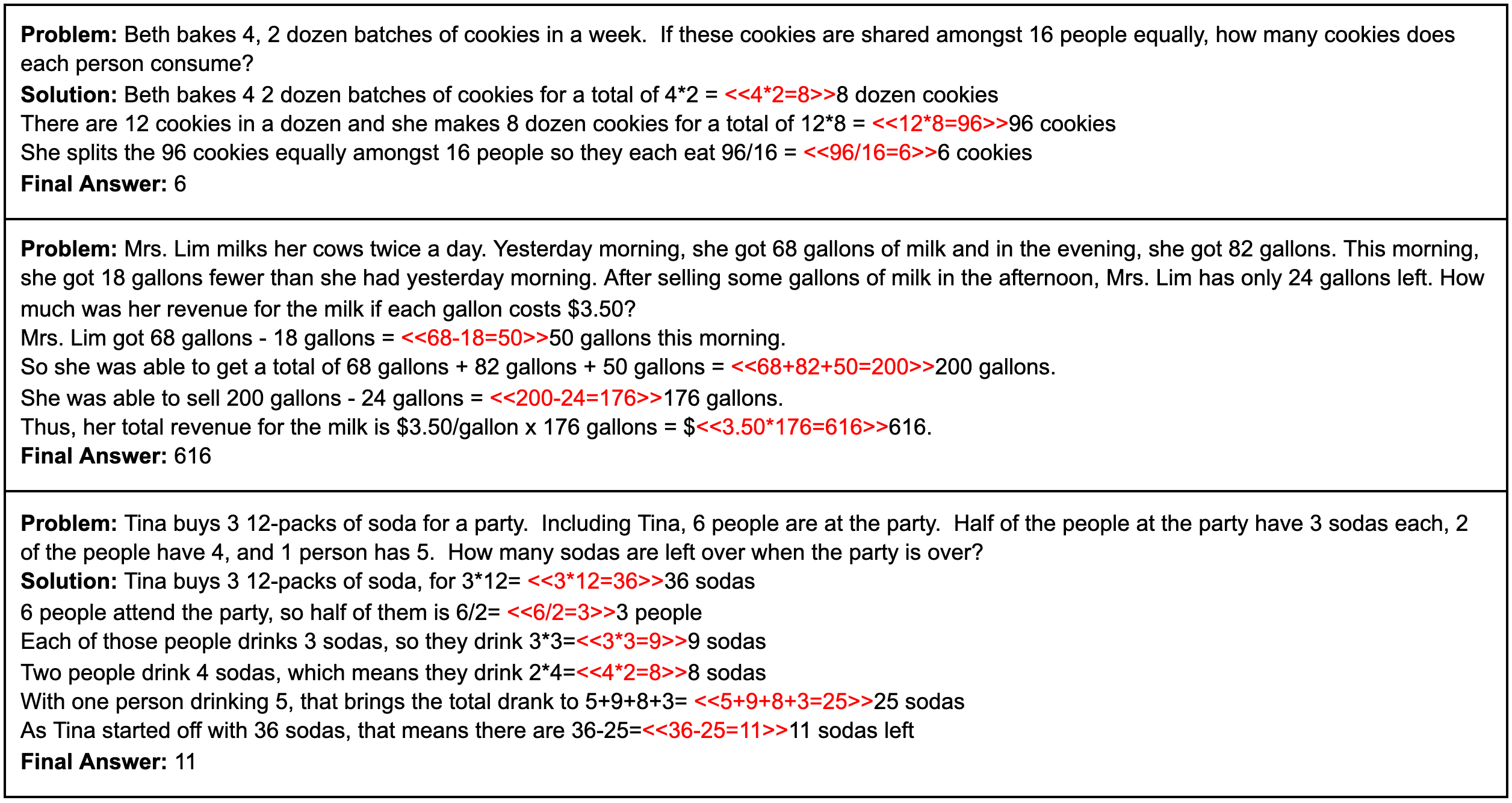}
\caption{Three example problems from GSM8K. Calculation annotations are highlighted in red.}
\label{fig:dataset_examples}
\end{figure*}

Our main contributions are as follows:

\begin{enumerate}
\item We present a curated dataset of 8.5K grade school math questions and natural language solutions, useful for probing the informal reasoning ability of large language models.

\item We show that, compared to a finetuning baseline, the use of verifiers results in approximately the same performance boost as a 30x model size increase, and that verifiers scale significantly better with increased data.

\item We show that dropout acts as a strong regularizer, significantly improving performance for both finetuning and verification.
\end{enumerate}

\section{Dataset} \label{section:dataset}

GSM8K consists of 8.5K high quality grade school math problems created by human problem writers. We segmented these into 7.5K training problems and 1K test problems. These problems take between 2 and 8 steps to solve, and solutions primarily involve performing a sequence of elementary calculations using basic arithmetic operations ($+ - \times \div$) to reach the final answer. A bright middle school student should be able to solve every problem.

We created GSM8K based on the following design principles.

\begin{itemize}
    \item \textbf{High Quality} We avoid error-prone scraping procedures and instead rely on human workers to create problems. After performing extensive quality control based on workers' answer agreement, we estimate that less than 2 percent of problems contain breaking errors.
    \item \textbf{High Diversity} We strive for high diversity among problems. We actively avoid designing problems that are drawn from the same linguistic template or differ only in superficial details, an issue that is prevalent among many other datasets. By creating each individual problem to be relatively unique, held-out test performance becomes a far more relevant metric.
    \item \textbf{Moderate Difficulty} We choose a problem distribution that is challenging for large state-of-the-art language models, without being completely intractable. GSM8K will help us better understand the data scaling trends of different models and methods in this difficulty sweet spot. Problems require no concepts beyond the level of early Algebra, and the vast majority of problems can be solved without explicitly defining a variable.
    \item \textbf{Natural Language Solutions} We collect solutions in natural language rather than as pure math expressions. We believe this is the most generally useful data format, and we expect it to shed light on the properties of large language models’ internal monologues. We instructed problem writers to explain their work as much as possible, but we allowed them to write solutions in their own diverse linguistic styles. 
\end{itemize}

The full GSM8K dataset can be found at \href{https://github.com/openai/grade-school-math}{https://github.com/openai/grade-school-math}. Example problems are shown in \Cref{fig:dataset_examples}, and we discuss additional dataset details in \Cref{appendix:dataset_details}.

\section{Related Work} \label{section:related}

\subsection{Related Datasets} 

Early math word problem datasets \citep{kushman2014learning, roy-roth-2015-solving} are relatively small and are not well suited for testing the limits of modern language models. Dolphin18K \citep{huang2016well} is a larger dataset containing 18K problems, but solutions are provided only in the form of equations or final answers. AQuA-RAT \citep{ling2017program} contains 100K problems, but this dataset unfortunately suffers from both a high degree of problem templatization and poor quality control of the natural language solutions. MathQA is a recently released subset of AQuA-RAT focused on correcting these mistakes \citep{amini2019mathqa}, but even the corrected dataset has data quality issues, with around 30\% of the data having inconsistencies \citep{miao2021diverse}. Ape210K \citep{zhao2020ape210k} is the largest publicly available dataset, consisting of 210K Chinese elementary school-level math problems. However, due to the language barrier and the lack of natural language solutions, we're unable to evaluate our methods on this dataset.

The recently developed ASDiv dataset \citep{miao2021diverse}, which contains 2.3K math word problems, addresses common flaws in prior datasets by ensuring problems have both high diversity and high quality. We share those design principles in the creation of GSM8K. However, we note that GSM8K is larger, provides natural language solutions, and consists of problems that on average require more steps to solve. The MATH dataset \citep{hendrycks2021measuring} is larger and significantly more complex than GSM8K, but the high difficulty makes it challenging to accurately measure progress given the current capabilities of state-of-the-art language models.

Other recent reasoning-related datasets have focused on mathematical reasoning on symbolic math \citep{lample2019deep}, reading comprehension (LogiQA) \citep{Liu2020LogiQAAC}, and commonsense question answering (CommonsenseQA) \citep{talmor2018commonsenseqa}. Similar to CommonsenseQA, GSM8K includes questions that require basic background knowledge, like the number of days in a week. Similar to LogiQA, which requires a mix of reading comprehension and logical reasoning, GSM8K's main difficulty lies in both properly interpreting a question and reasoning through the steps to solve it.

\subsection{Related Methods}

Previous work has attempted to solve classic math word problem benchmarks with recurrent seq2seq models \citep{sutskever2014sequence} and closely related variants \citep{wang-etal-2017-deep, huang2018neural}. More recent work has improved performance by designing specialized encoder-decoder architectures \citep{amini2019mathqa, chiang2018semantically, Xie2019AGT, Chen2020MappingNP, li2020graph}, with the strongest results often relying on large pretrained encoders from the BERT family \citep{chen2019neural, kim2020point, mwpbert}.

Other recent work has recommended additional pretraining tasks to further improve the math reasoning skills of large transformer-based models. \cite{hendrycks2021measuring} propose pretraining models on a new AMPS corpus, derived from Khan Academy problems and Mathematica scripts. Similarly, \cite{shen2021mathbert} propose a pretrained a corpus of pre-K to college level curricula extracted from the internet, and \cite{Peng2021MathBERTAP} propose pretraining by predicting masked subexpressions from expression trees.

Similar to verification, other methods have finetuned a language model to select among many model completions. \cite{nichols2020collaborative} proposed a sample-and-rank approach to improve the collaborative storytelling ability of large language models, with the training signal coming from the preferences of human workers. In concurrent work closely related to our own, \cite{shen2021generate} applied a similar approach to solving math word problems, jointly training a model to both generate and rank solutions. Our work shares many fundamental similarities with their approach, though we differ in several key respects. First, we focus attention on the space of natural language solutions, as this is a richer and more general solution format than pure mathematical expressions. Moreover, this choice enables our models to develop verbal analytical skills and to produce solutions that are more readily interpretable by humans. Second, we provide evidence that verifiers scale far more favorably with additional data than baseline methods. Finally, we use separate generator and verifier networks, in order to prevent the generator from overfitting.

\section{Methods}

\begin{figure*}
\centering
\includegraphics[width=.475 \textwidth]{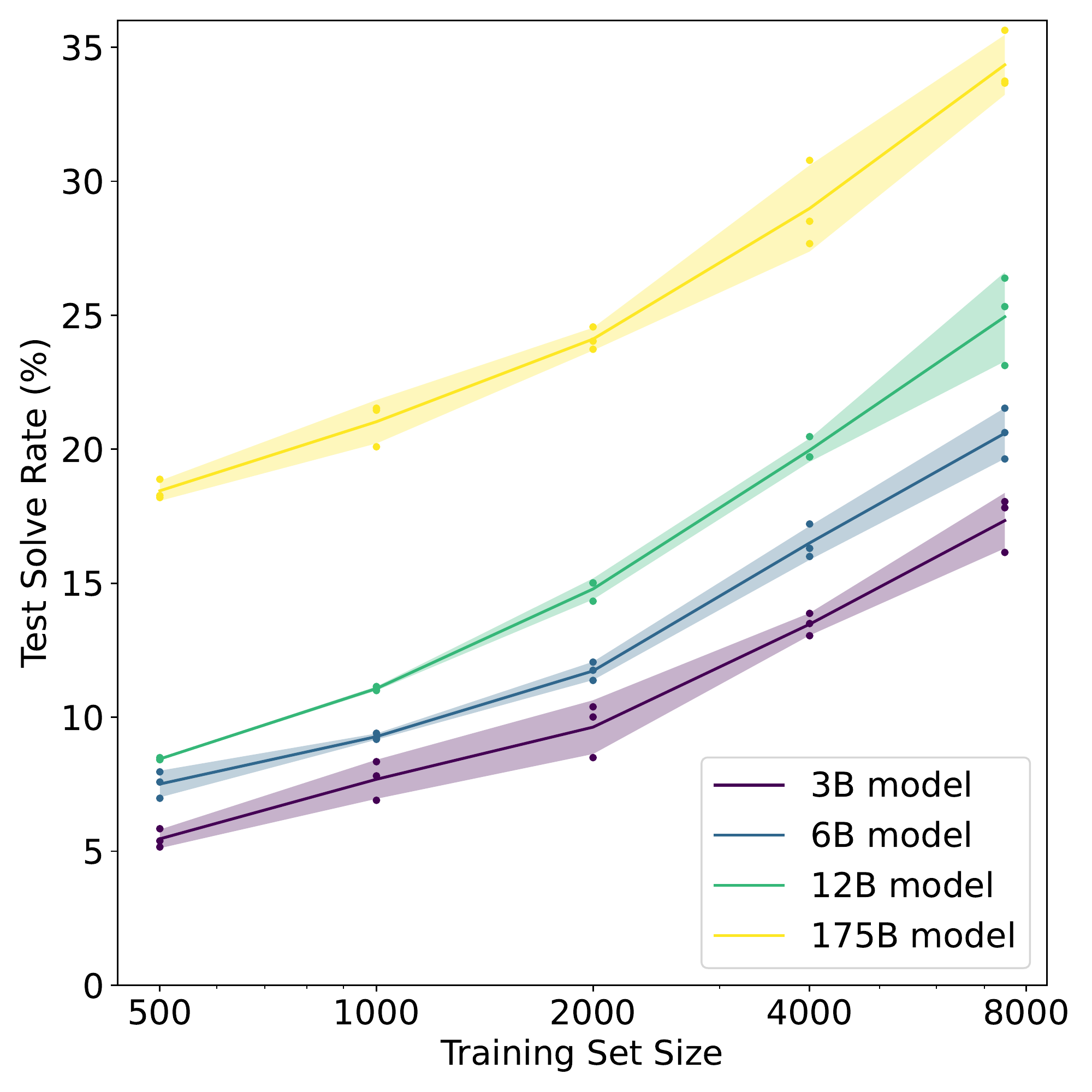}
\includegraphics[width=.475 \textwidth]{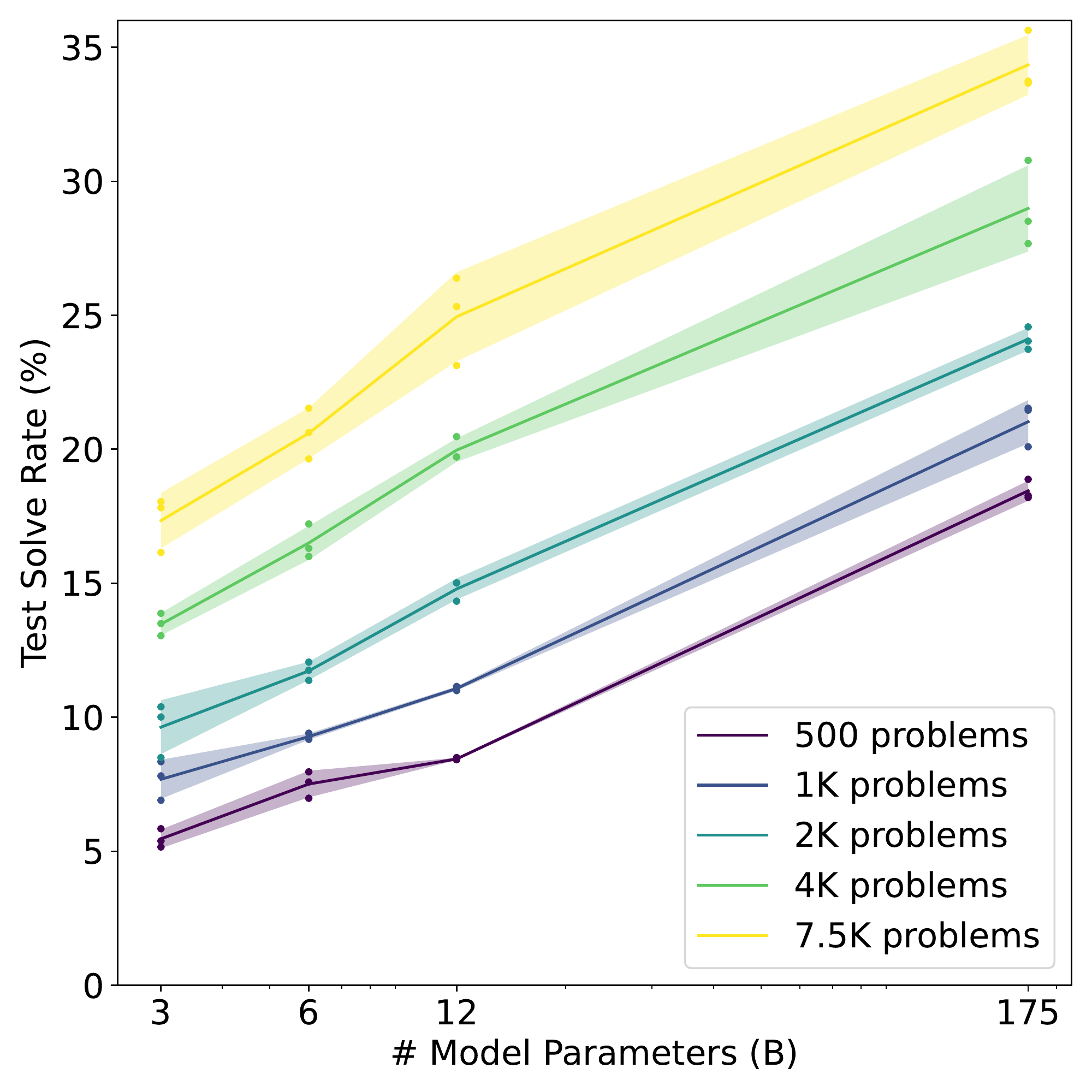}
\caption{Final test performance for various GPT-3 model sizes after finetuning on training sets of different sizes. Mean and standard deviation is shown across 3 runs.}
\label{fig:bc_baseline}
\end{figure*}

We investigate two methods to solve problems in GSM8K: finetuning and verification. Finetuning, our baseline method, uses the same language modeling objective as the generative pretraining in GPT-3 \citep{brown2020language}. At test time, we judge performance by autoregressively sampling a single low temperature solution and checking whether the final answer is correct. In contrast, verification consists of sampling multiple high temperature solutions, assigning each solution a score, and outputting the highest ranked solution. Verifiers are trained to judge the correctness of solutions, with the training signal determined solely by whether or not the solution reached the correct final answer.

For both methods, we use models from the GPT-3 family as our initialization, primarily focusing on the 175B and 6B model sizes. The 175B model is the largest and produces the most impressive results, while the 6B model is significantly more convenient for research purposes. We discuss hyperparameter choices in \Cref{appendix:hyperparameters}.

Our models frequently fail to accurately perform calculations. Although larger models make fewer arithmetic mistakes than smaller models, this remains a common source of errors. To mitigate this issue, we train all models to use a calculator by injecting calculation annotations into the training set. At test time, a calculator will override sampling when the model chooses to use these annotations. Details can be found in \Cref{appendix:calculator_annotations}.

\begin{figure*}
\centering
\includegraphics[width=.475 \textwidth]{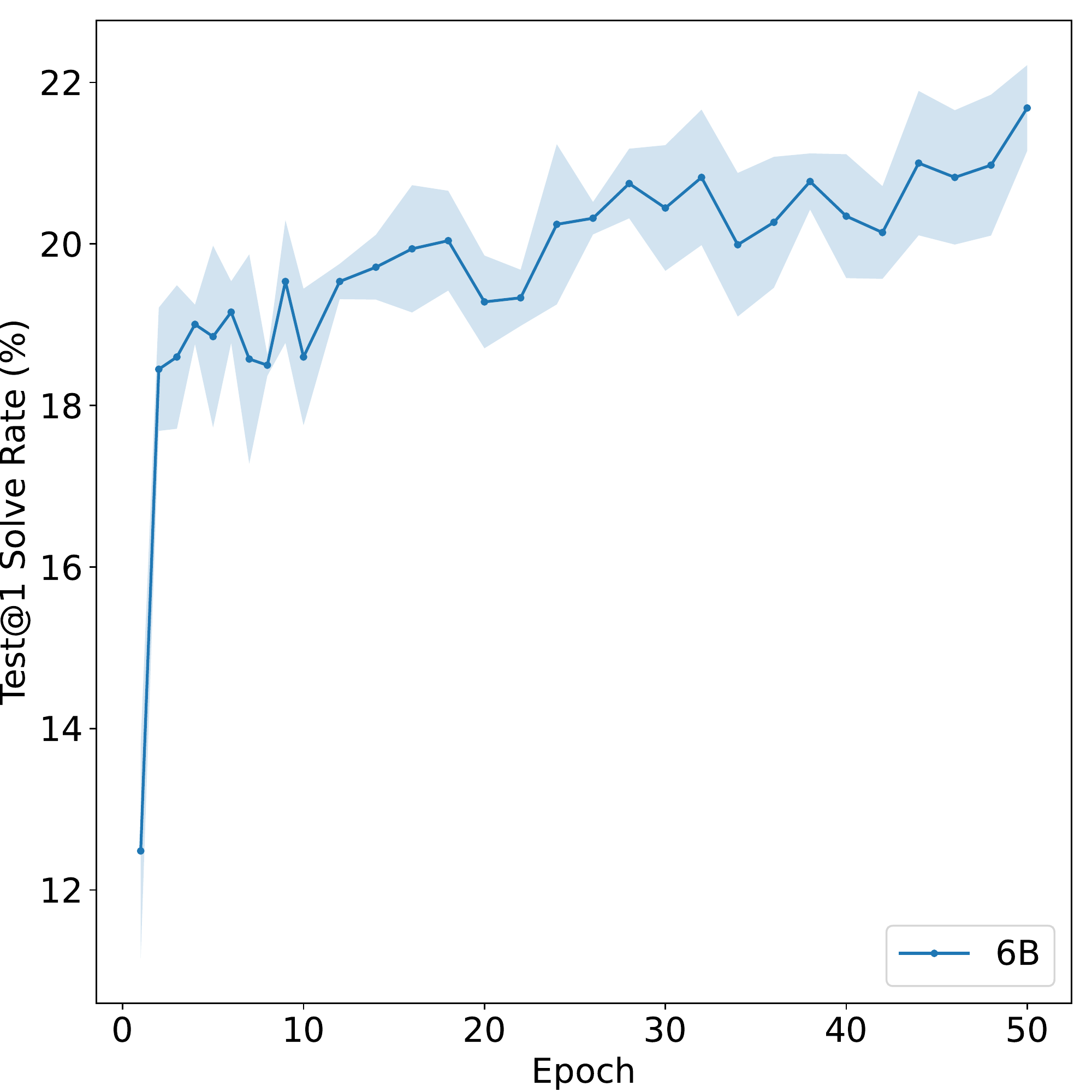}
\includegraphics[width=.475 \textwidth]{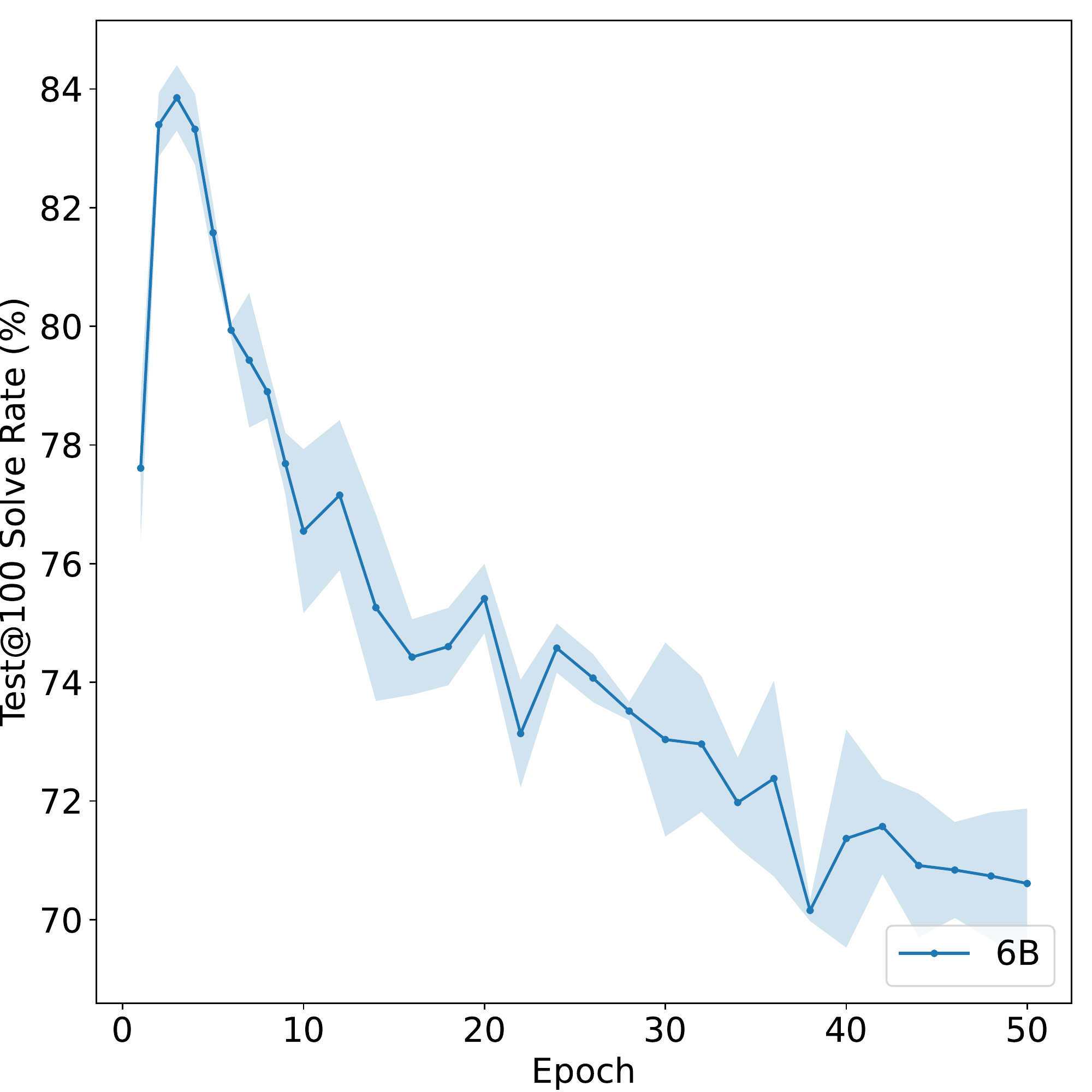}
\caption{Test solve rate after finetuning a 6B model on the full GSM8K training set, when the model is allowed to make 1 guess (left) or 100 guesses (right).}
\label{fig:bc_training}
\end{figure*}

\subsection{Finetuning}

We perform finetuning by updating model parameters to minimize the cross-entropy loss over all training tokens. \Cref{fig:bc_baseline} shows test performance after finetuning on training sets of varying sizes for 20 epochs. We visualize the same data both as a function of training set size and as a function of model size. Test performance is determined by a single low temperature ($T=0$) sample for each test problem. Unsurprisingly, we see that the 175B model significantly outperforms the smaller models. Assuming a log-linear trend, we can naively extrapolate these results to estimate that a model with $10^{16}$ parameters would be required to reach an $80\%$ solve rate, when using the full GSM8K training set. It is even harder to extrapolate along the data dimension, since performance does not appear to follow a log-linear trend. Nevertheless, it appears likely that the 175B model would require at least two additional orders of magnitude of training data to reach an $80\%$ solve rate.

In \Cref{fig:bc_training}, we show how 6B test performance varies over the course of 100 training epochs. We use test@N to denote the percentage of problems solved correctly at least once when allowing the model to make N separate guesses for each problem. We use a low temperature ($T=0$) to generate test@1 samples and we use a higher temperature ($T=0.7$) to generate test@100 samples. Both temperature values were chosen empirically to produce the best results. Test@1 performance improves approximately monotonically, even though we quickly begin overfitting on test loss. Unfortunately, test@100 performance degrades much more sharply than test@1 as we increase the number of epochs. This is to be expected: as the model repeatedly encounters the same data, it becomes increasingly uncalibrated and overconfident in its predictions. At test time, this overconfidence leads to poor coverage of the solution space, an effect which only becomes noticeable when we are considering multiple samples at test time.

Choosing a model with good coverage is critical to successfully train verifiers. Empirically, we see that test@100 performance peaks within the first few epochs. For this reason, we use models trained for 2 epochs to generate samples for training verifiers. We provide several example solutions from 6B and 175B models in \Cref{appendix:example_solutions}. We also note that it is important to allow the model to generate the full natural language solution before outputting a final answer. If we instead finetune a 6B model to directly output the final answer without any intermediate steps, performance drops drastically from 20.6\% to 5.2\%.

\begin{figure}
\centering
\begin{subfigure}{1.0 \textwidth}
\includegraphics[width=\textwidth]{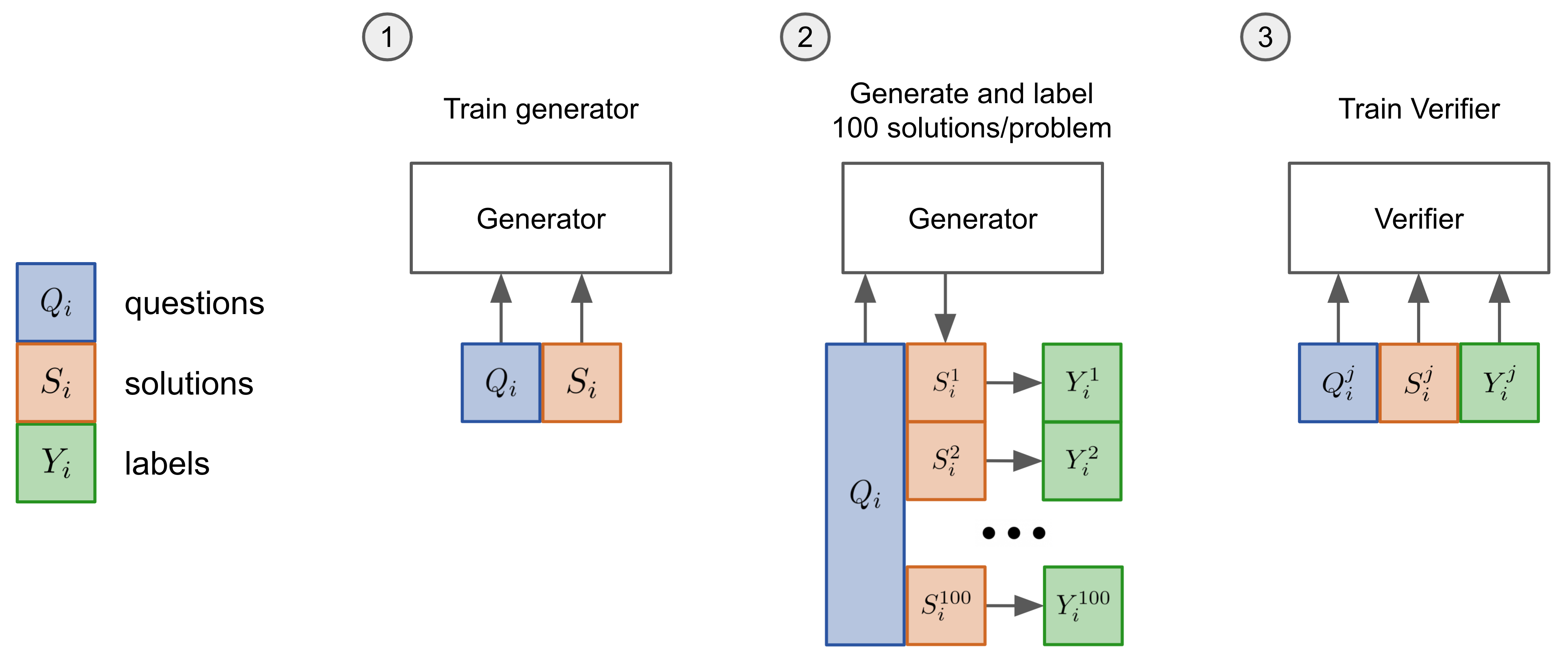}
\end{subfigure}
\caption{A diagram of the verification training pipeline.}
\label{fig:verifier_diagram}
\end{figure}

\subsection{Verification} \label{section:verification}

To improve upon the finetuning baseline, we train verifiers to judge the correctness of model-generated solutions and search against these verifiers at test time. Conditioned on the problem and a candidate solution, the verifier outputs the probability that the solution is correct. Training solutions are labeled as correct or incorrect based solely on whether they reach the correct final answer. In practice, some solutions will reach the correct final answer using flawed reasoning, leading to false positives.

As shown in \Cref{fig:verifier_diagram}, we train the verifier as follows:

\begin{enumerate}
\item Finetune a model (the “generator”) for 2 epochs on the training set.
\item Sample 100 completions from the generator for each training problem and label each solution as correct or incorrect.
\item Train a verifier for a single epoch on this dataset.
\end{enumerate}

Training for 2 epochs is enough for the generator to learn basic skills in this domain. We choose not to train for longer, since the diversity of generated solutions begins to collapse after this point, as shown in \Cref{fig:bc_training}. We train separate generator and verifier models to limit the generator's training and prevent overfitting, but in principle, it should be possible to combine these models. Unless otherwise specified, we use the same model size for the generator and the verifier. In addition to predicting solution correctness, we also train the verifier with the same language modeling objective as the generator. This serves as a valuable auxiliary objective for the verifier. We discuss additional verifier training details in \Cref{appendix:verifier_details}.

\begin{figure}
\centering
\begin{subfigure}{0.475 \textwidth}
\includegraphics[width=\textwidth]{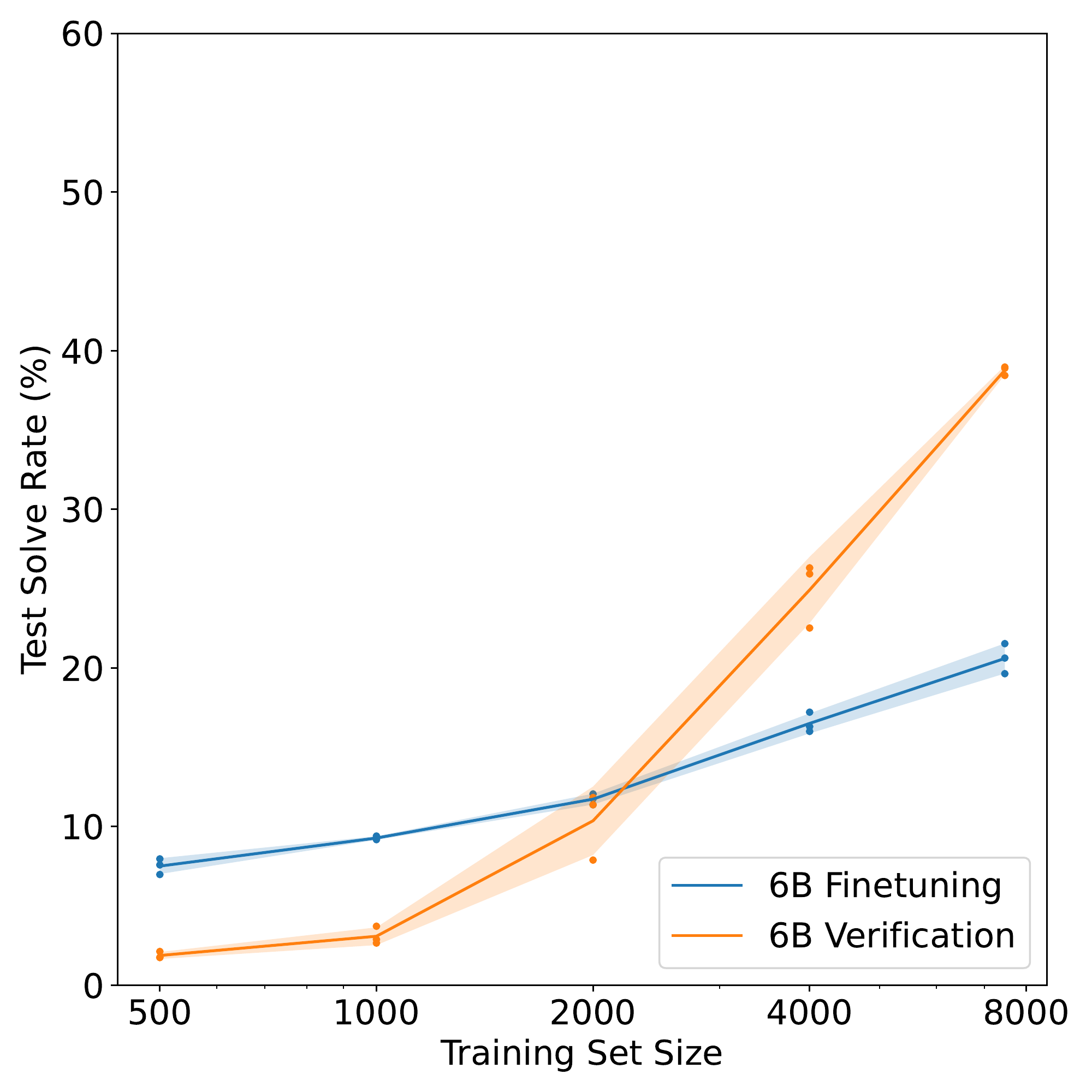}
\end{subfigure}
\hspace*{\fill}
\begin{subfigure}{0.475 \textwidth}
\includegraphics[width=\textwidth]{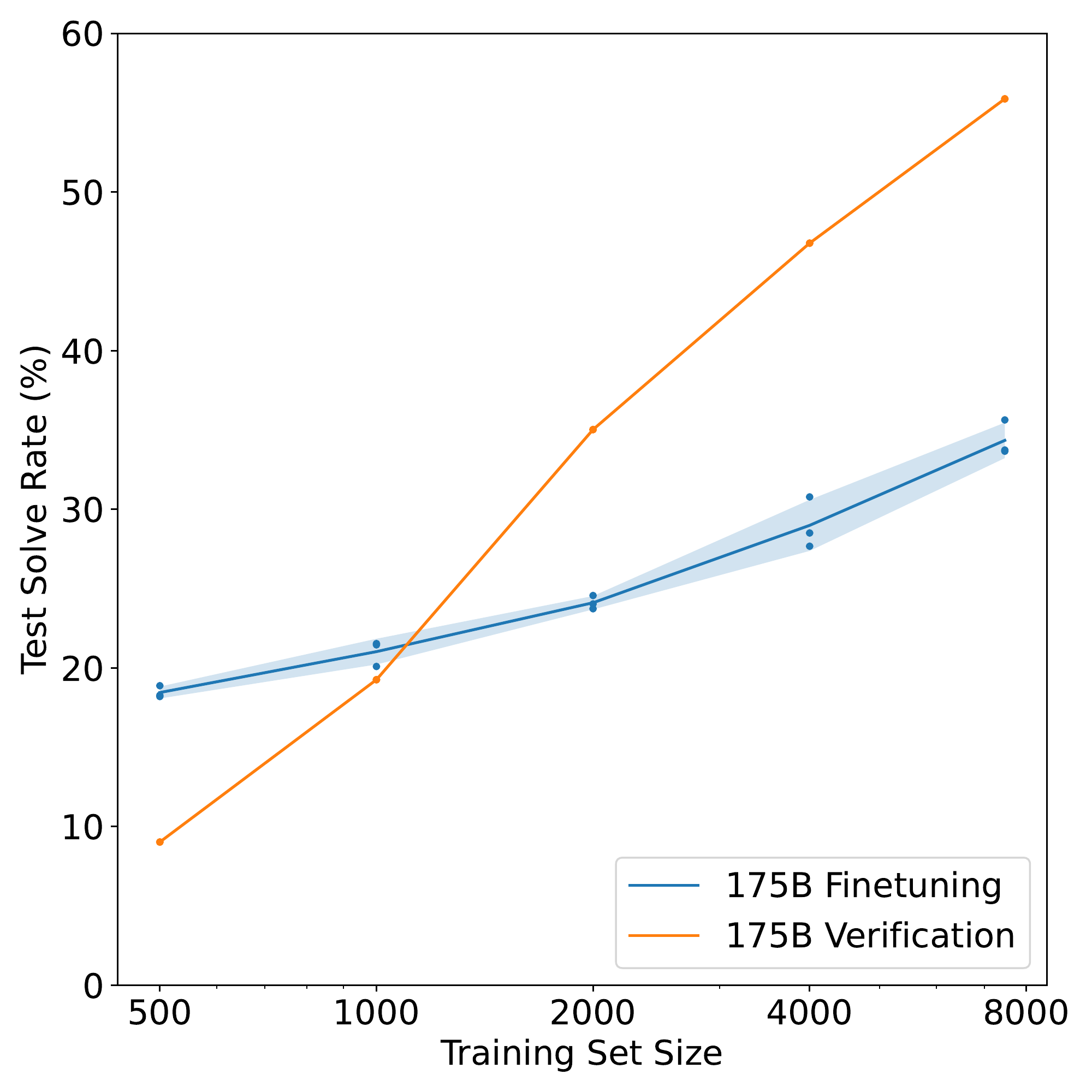}
\end{subfigure}
\caption{A comparison between finetuning and verification using 6B and 175B model sizes. Verification considers 100 solutions per problem. Mean and standard deviation is shown across 3 runs, except for 175B verification which shows only a single run.}
\label{fig:full_completion_vf}
\end{figure}

At test time, we sample 100 completions to each test problem, rank them with the verifier, and then return the one with the highest verifier score. A comparison between verification and finetuning is shown in \Cref{fig:full_completion_vf} for both the 6B and 175B model sizes. We find that it is not beneficial to use verification at low dataset sizes. We believe this is due to the pressure to overfit to the correct answer: with small datasets, overfitting to the correct answer happens faster than learning more generalizable properties of correct reasoning. However, once we use a sufficiently large dataset, we see a strong boost from verifiers. It’s interesting to note that the 175B verifiers “take off” earlier than the 6B verifiers, requiring fewer training problems to surpass the finetuning baseline. See \Cref{appendix:example_solutions} for example solutions found by verifiers and \Cref{appendix:verifier_visualization} for a visualization of verifier confidence. 

\begin{figure}
\centering
\begin{subfigure}{0.32 \textwidth}
\includegraphics[width=\textwidth]{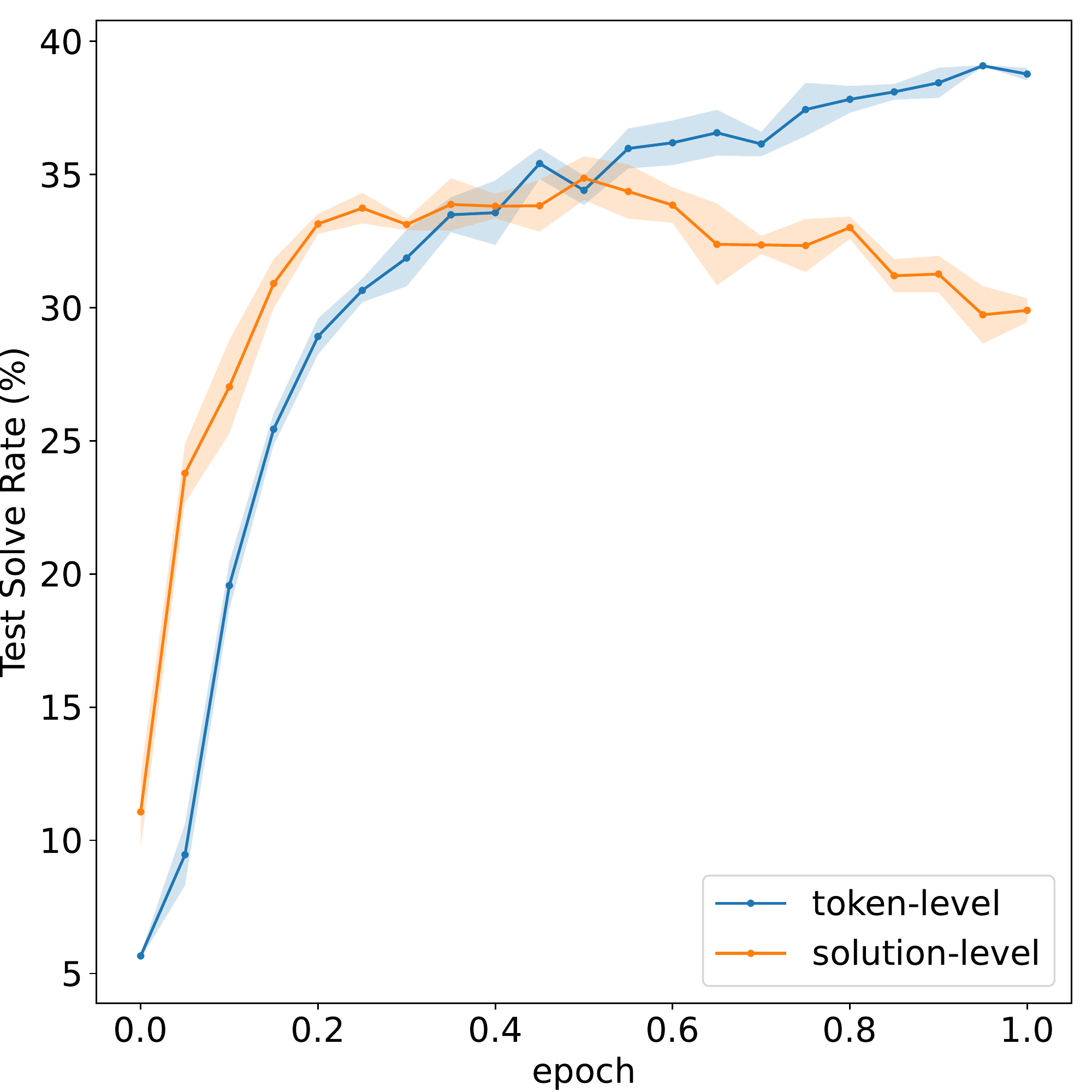}
\caption{Comparison between a verifier trained to predict correctness after every token (token-level) and one trained to predict correctness after only the final token (solution-level)} \label{fig:fc_verifier_token_ablation}
\end{subfigure}
\hspace*{\fill}
\begin{subfigure}{0.32 \textwidth}
\includegraphics[width=\textwidth]{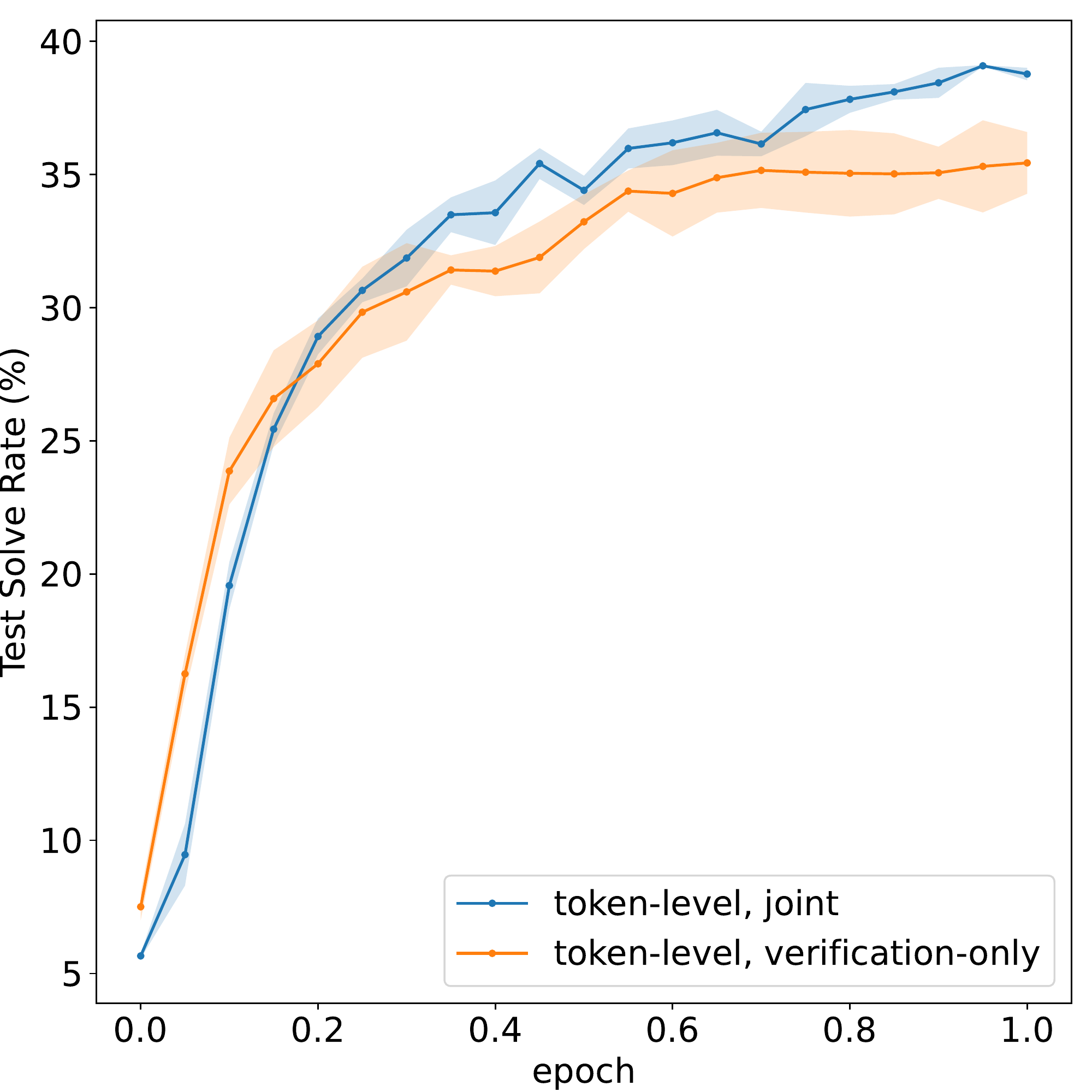}
\caption{Comparison between a verifier trained jointly to predict correctness and perform language modeling (joint) and one trained only to predict correctness (verification-only)} \label{fig:fc_verifier_loss_ablation}
\end{subfigure}
\hspace*{\fill}
\begin{subfigure}{0.32 \textwidth}
\includegraphics[width=\textwidth]{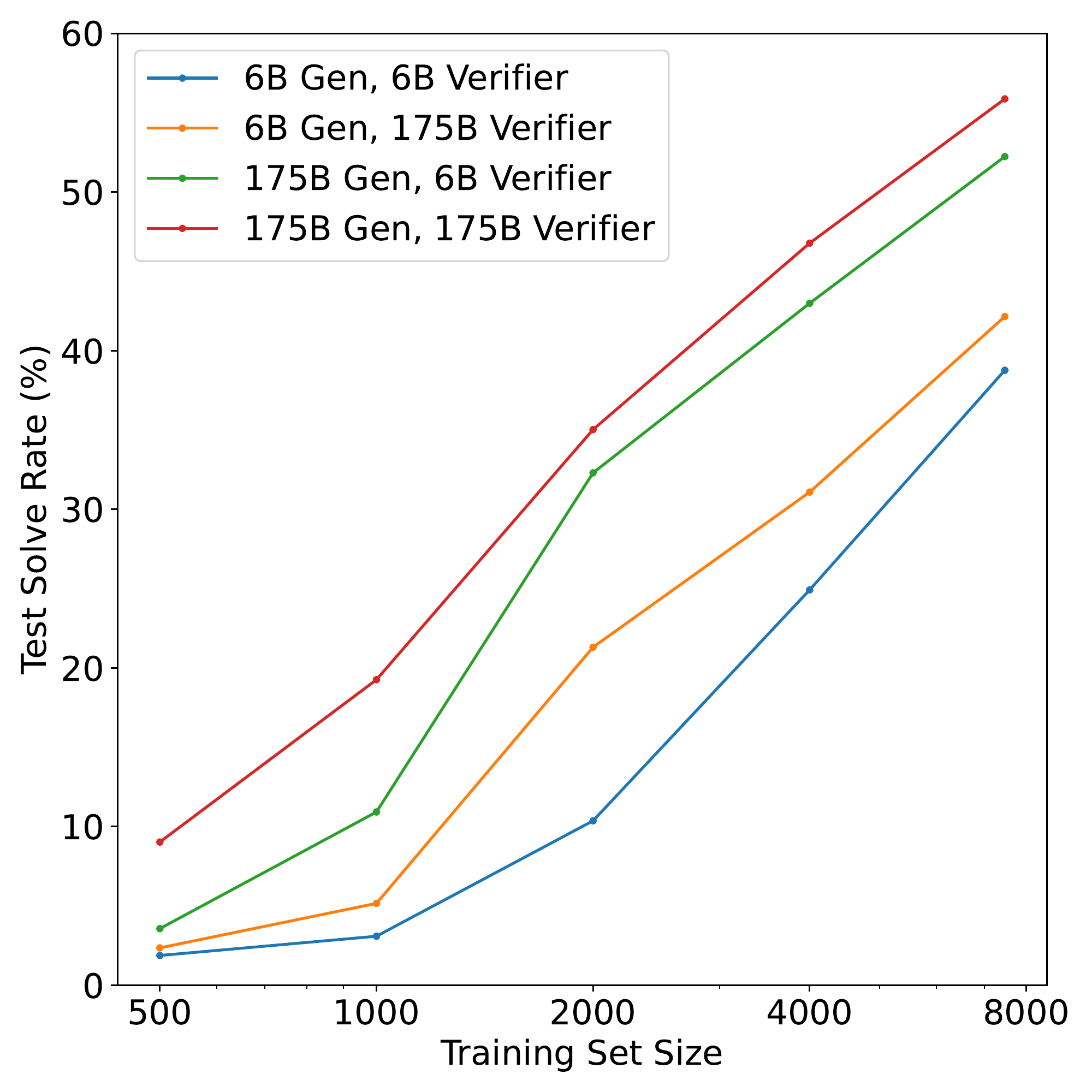}
\caption{Performance when varying the size of the generator and the verifier in isolation. Increasing the size of the generator has a larger impact than increasing the size of the verifier.} \label{fig:gen_vf_ablate_size}
\end{subfigure}
\caption{Verification ablations}
\label{fig:fc_verifier_ablations}
\end{figure}

\subsection{Verification Ablations} \label{section:verifier_ablations}

We can either train verifiers to make a single scalar prediction conditioned on the entire generated solution, or to make a scalar prediction after each token in the solution. By default, we choose the latter, training verifiers to make predictions after each token. This can be viewed as a token-level value function. We compare these two methods in \Cref{fig:fc_verifier_token_ablation}, respectively labeled “solution-level” and “token-level”.

Predicting the value function at every token is a more challenging and noisier task than judging only the full completion. However, despite the initially slower training, the token-level verifier ultimately outperforms the solution-level verifier. Moreover, the token-level verifier is still improving late in training, whereas the solution-level verifier quickly shows signs of overfitting. We hypothesize that the full value function provides a useful auxiliary signal that encourages the model to judge the reasoning throughout solutions, rather than merely memorizing the correct final answer. 

In \Cref{fig:fc_verifier_loss_ablation}, we ablate the objective used when training verifiers. As discussed in \Cref{section:verification}, we can optionally include a language modeling objective alongside the verification objective. We compare using both objectives to using only the verification objective. Although both are reasonable choices, including the language modeling objective is a strict improvement. This makes intuitive sense: better understanding this language distribution should only aid the verifier in discriminating between samples.

In \Cref{fig:gen_vf_ablate_size}, we separately ablate the model size of the generator and the verifier. We find that using a large generator with a small verifier performs significantly better than using a small generator with a large verifier. Verification is still remarkably effective, even when the verifier is much smaller than the generator. This suggests that the verifier may often be relying on relatively coarse heuristics to discriminate between solutions from a given generator, rather than attempting a more thorough form of verification.

\begin{figure}
\centering
\begin{subfigure}{0.475 \textwidth}
\includegraphics[width=\textwidth]{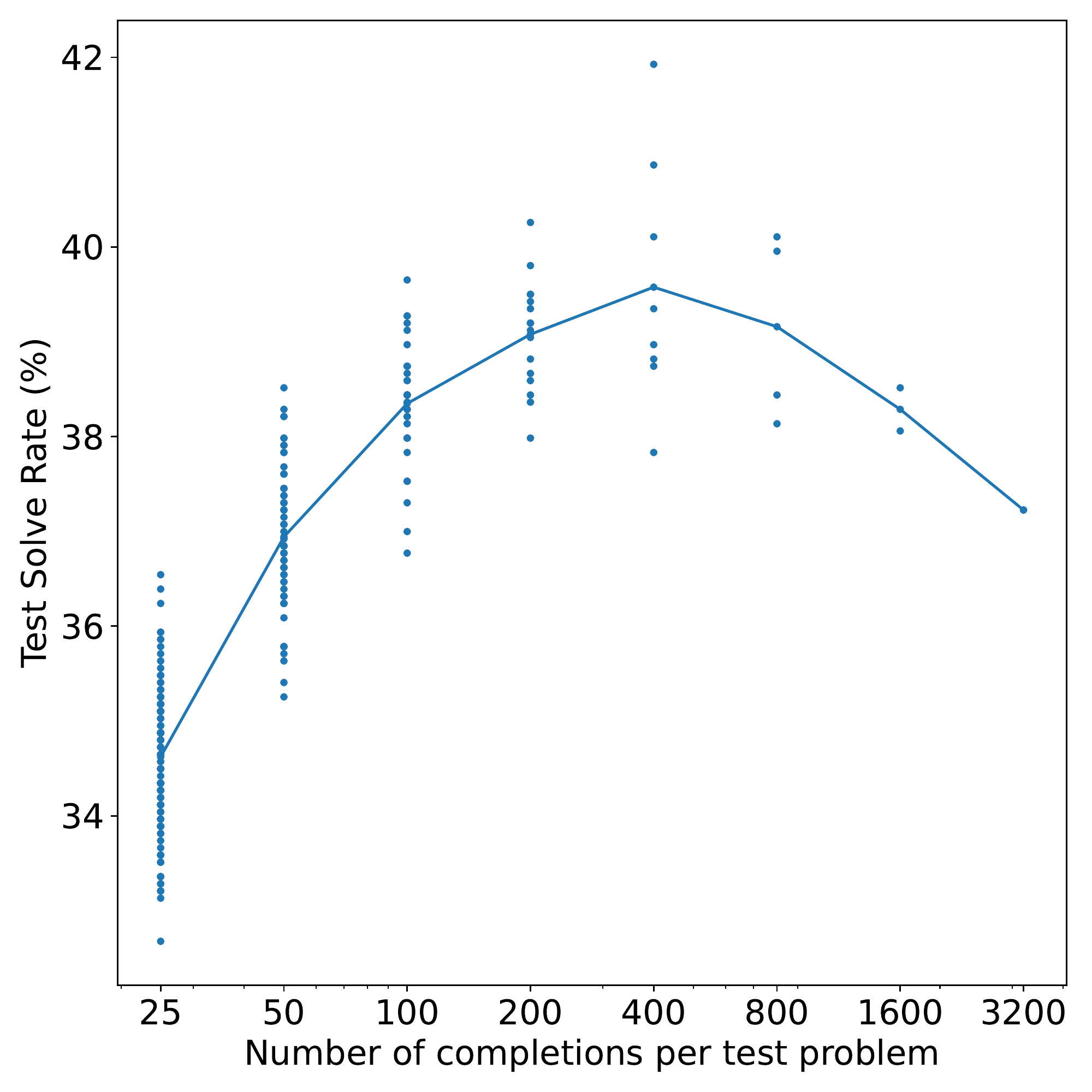}
\caption{6B verification test performance when given varying numbers of completions per problem to rank.} \label{fig:completions}
\end{subfigure}
\hspace*{\fill}
\begin{subfigure}{0.475 \textwidth}
\includegraphics[width=\textwidth]{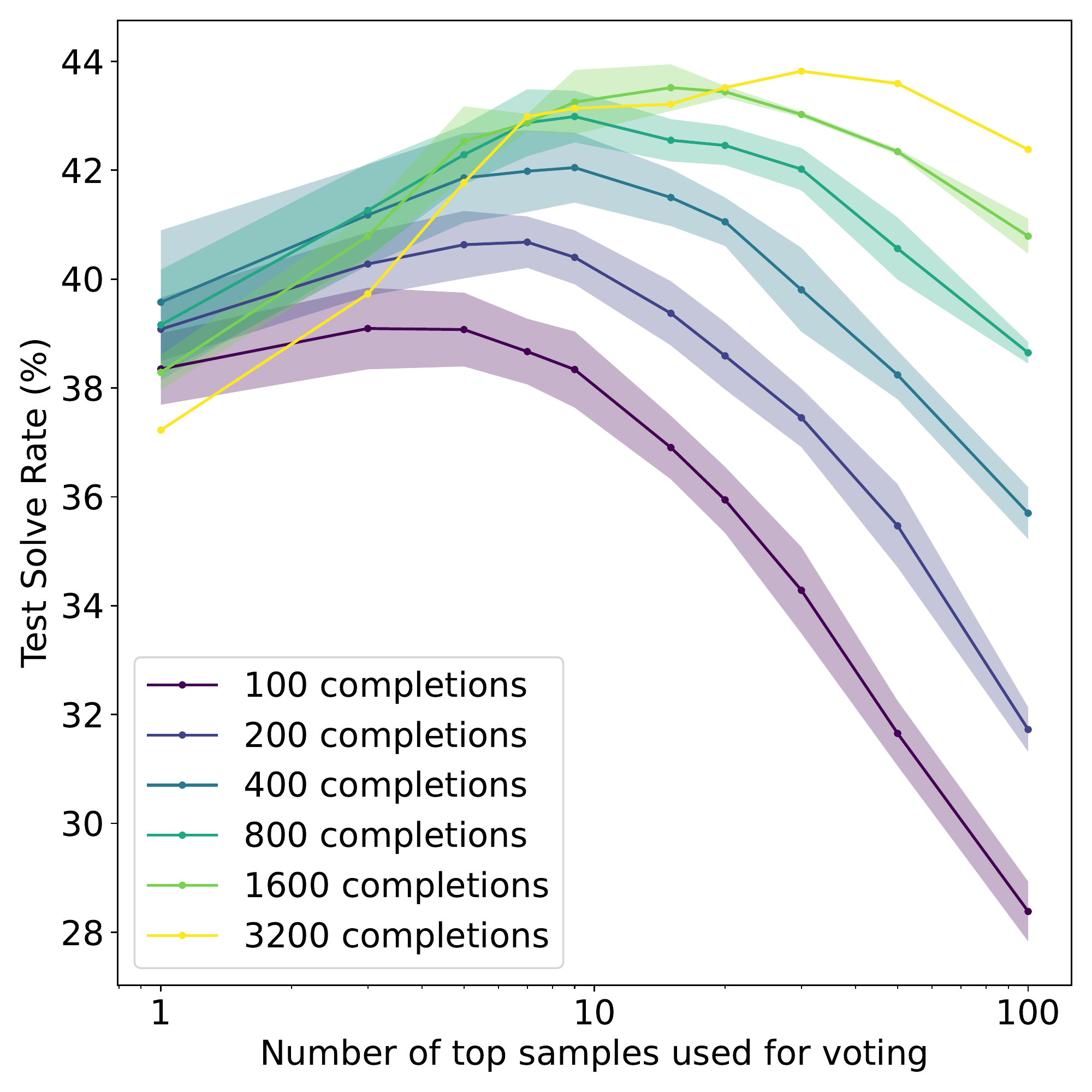}
\caption{6B verification test performance when varying the number of top ranked samples allowed to vote on the answer.} \label{fig:vote}
\end{subfigure}
\caption{Performance as the amount of test time compute varies.}
\label{fig:test_time_compute_trends}
\end{figure}

\section{Additional Experiments}

\subsection{Test Time Compute}

At test time, we can choose to generate arbitrarily many solutions to be judged by the verifier before selecting the highest ranked completion. \Cref{fig:completions} shows how 6B verifier performance varies with the number of completions per test problem. At this scale, performance improves as we increase the number of completions up to 400. Beyond this point, performance start to decrease. This suggests that the benefits of search are eventually outweighed by the risk of finding adversarial solutions that fool the verifier. In general, we evaluate verifier test performance using 100 completions, since this captures most of the benefits of verification with a relatively modest compute cost.

To further increase performance, we can take a majority vote among the top verifier-ranked solutions instead of selecting only the single top solution. This voting process considers only the final answer reached by the individual solutions: the final answer selected is the one with the most votes. \Cref{fig:vote} shows how performance varies as we allow a greater number of top samples to cast a vote. Unsurprisingly, when starting with a greater number of samples, we can afford to allow a greater number of samples to cast a vote. When we have only 100 samples, it is optimal to allow only the top 3-5 samples to cast a vote. When we have 3200 samples, it is approximately optimal to allow the top 30 to cast a vote.

\subsection{Regularization}

\begin{figure}
\centering
\begin{subfigure}{0.32 \textwidth}
\includegraphics[width=\textwidth]{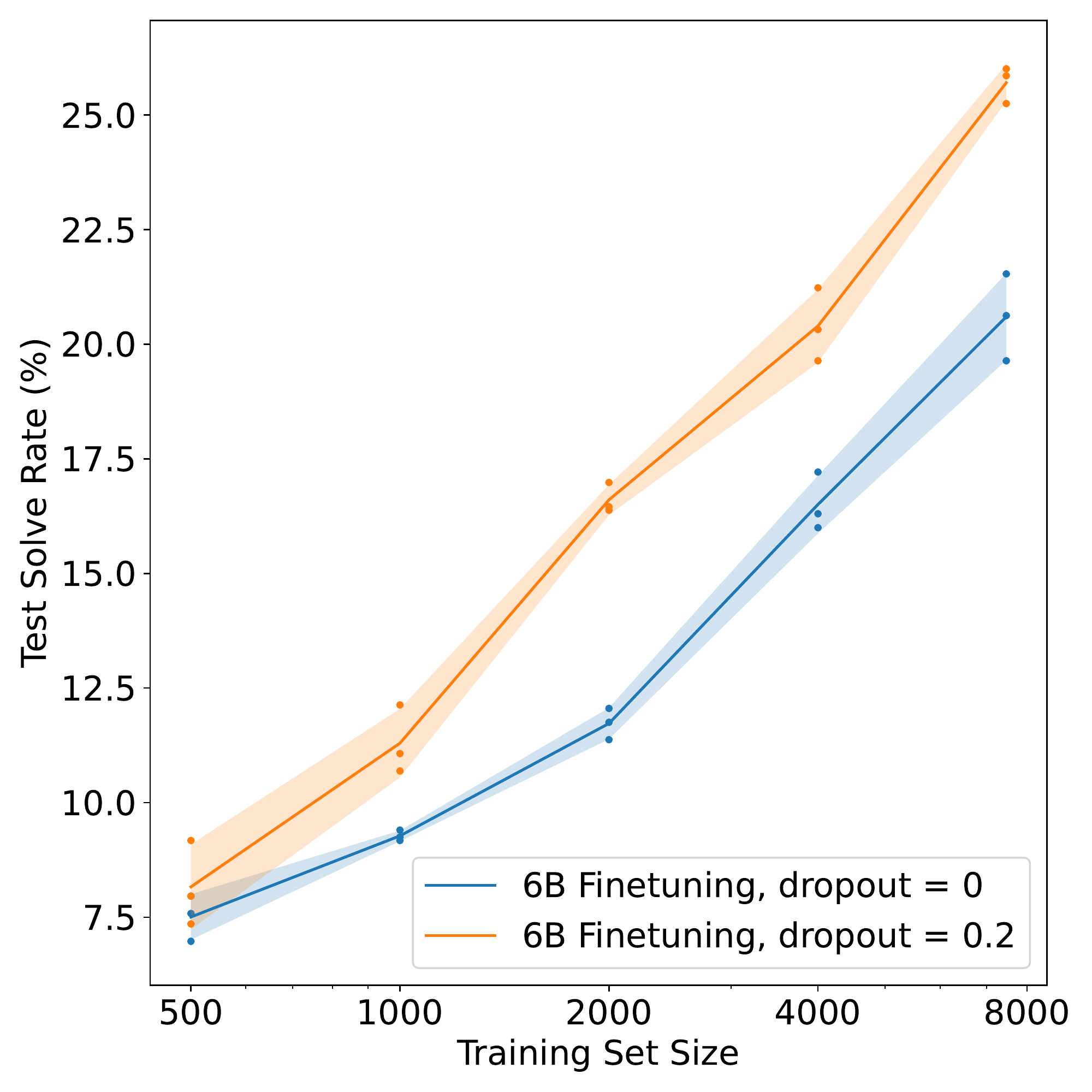}
\caption{Finetuning} \label{fig:bc_dropout}
\end{subfigure}
\hspace*{\fill}
\begin{subfigure}{0.32 \textwidth}
\includegraphics[width=\textwidth]{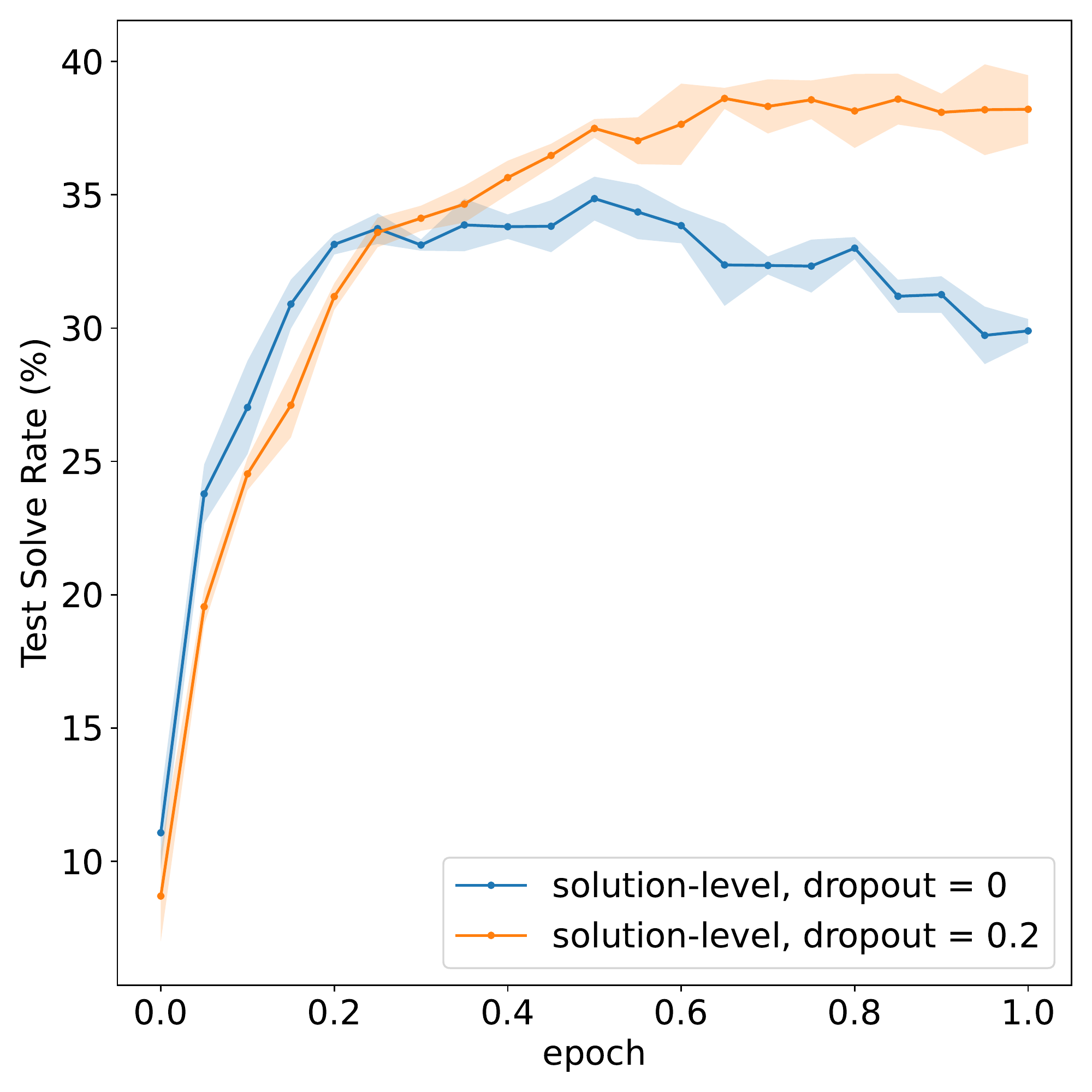}
\caption{Solution-level verifiers} \label{fig:single_token_dropout}
\end{subfigure}
\hspace*{\fill}
\begin{subfigure}{0.32 \textwidth}
\includegraphics[width=\textwidth]{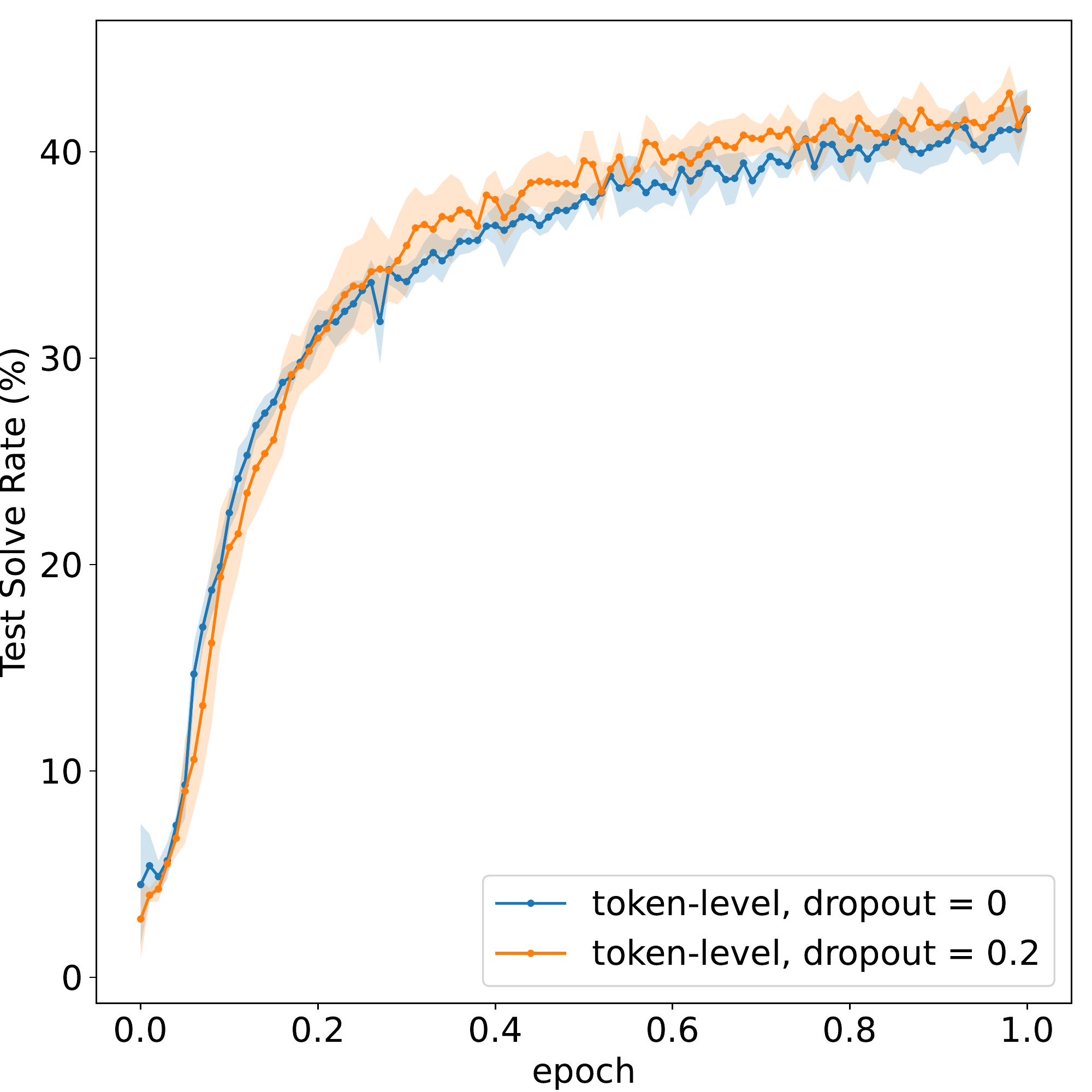}
\caption{Token-level verifiers} \label{fig:all_token_dropout}
\end{subfigure}
\caption{6B finetuning and verification dropout ablations.}
\label{fig:dropout_ablations}
\end{figure}

 We find that both finetuning and verification strongly benefit from the use of dropout as a regularizer. Specifically, we apply residual dropout \citep{vaswani2017attention} along the residual paths of each layer in the network. We use 20\% dropout for all dropout experiments, chosen based on the results of a hyperparameters sweep. We note that GPT-3 models are not pretrained with dropout. For experiments involving dropout, we therefore perform additional pretraining with dropout before subsequently finetuning the models. This mitigates the distribution shift the model experiences during finetuning.

 We first investigate the effect of dropout on finetuning across various training set sizes. \Cref{fig:bc_dropout} shows that dropout leads to a significant improvement over baseline. We next investigate the effect of dropout on verifiers, considering both the solution-level and token-level variants. In \Cref{fig:single_token_dropout}, we see that dropout significantly improves solution-level verifiers, mitigating the overfitting that occurs in the unregularized baseline. Notably, using dropout with solution-level verifiers reaches a similar level of performance as token-level verifiers. In \Cref{fig:all_token_dropout}, we apply dropout to token-level verifiers. Since token-level verifiers are already less susceptible to overfitting, it is no surprise that the impact of dropout is less significant. Nevertheless, we do still see a slight gain from training token-level verifiers with dropout. Note that we increase the batch size for token-level verifiers by a factor of 4, to better handle the more difficult objective and the noise from dropout.

\section{Conclusion}

We have seen that verification provides a significant performance boost relative to a finetuning baseline. On the full dataset, 6B verification slightly outperforms a finetuned 175B model, thereby offering a boost approximately equivalent to a 30x model size increase. We have also seen that token-level verifiers are less prone to overfitting than solution-level verifiers, and that all methods benefit from regularization with residual dropout. We expect verification to scale well to problem distributions that require more complex mathematical reasoning, and we hope GSM8K supports the development of new methods that scale even better.

\section*{Acknowledgements}

We thank Dan Hendrycks, Leo Gao, Alec Radford, and Giambattista Parascandolo for their valuable feedback on this paper; Harri Edwards, Yura Burda, Michael Wu, and Nick Ryder for many insightful conversations; Michael Petrov, Alethea Power, and Jacob Jackson for their technical assistance; the OpenAI Supercomputing team for the infrastructure that made these experiments possible; and the team at Surge AI for performing the GSM8K data collection.

\changeurlcolor{black}

\bibliography{grade_school_math}

\changeurlcolor{blue}

\appendix

\clearpage

\section{Dataset Details} \label{appendix:dataset_details}

We initially collected a starting set of a thousand problems and natural language solutions by hiring freelance contractors on Upwork (\url{upwork.com}). We then worked with Surge AI (\url{surgehq.ai}), an NLP data labeling platform, to scale up our data collection. After collecting the full dataset, we asked workers to re-solve all problems, with no workers re-solving problems they originally wrote. We checked whether their final answers agreed with the original solutions, and any problems that produced disagreements were either repaired or discarded. We then performed another round of agreement checks on a smaller subset of problems, finding that 1.7\% of problems still produce disagreements among contractors. We estimate this to be the fraction of problems that contain breaking errors or ambiguities. It is possible that a larger percentage of problems contain subtle errors.

To assist contractors with writing questions, we provided seed questions automatically generated from a few-shot prompted 175B GPT-3 model. Contractors were allowed to use those seed questions directly, to use them as inspiration and make modifications, or to come up with their own questions entirely. We instructed contractors to be as descriptive as possible in their solutions, and to not re-use problem settings or templates between different questions. To ensure contractors were not re-using problem templates, we computed pairwise similarity scores between problems and used this to provide feedback to contractors.

\newpage

\section{Hyperparameters} \label{appendix:hyperparameters}

We include a table of important hyperparameters below. We performed sweeps of the learning rate and batch size by an order of magnitude in both directions from the values in the table and were unable to find any significant improvements. Other reasonable choices for both the verifier temperature (eg: 1.0 instead of 0.7) and objective (cross-entropy instead of mean squared error) also had negligible effect in our ablations.

\begin{table}[h!]
    \footnotesize
    \centering
    \renewcommand{\arraystretch}{1.3}
    \begin{tabular}{@{}ll@{}}
        \toprule
        \textbf{General Hyperparameters} & \textbf{Value} \\ 
        \midrule
        Batch Size & $3.2 \times 10^4$ tokens \\
        Max Sample Length & 400 tokens \\
        Tokenization & reversible\_50000 \\
        Optimizer & Adam, $\beta_1 = 0.9$, $\beta_2 = 0.95$ \\
        Dropout & $0.0$ \\
        Learning Rate Schedule & Linear decay to 0 \\
        \midrule
        \textbf{Finetuning Hyperparameters} & \textbf{Value} \\ 
        \midrule
        Epochs & 20 \\
        Sampling Temperature & 0 (argmax) \\
        Base Learning Rate ($\alpha$) & $1.6 \times 10^{-5}$ (3B) \\
        & $1.2 \times 10^{-5}$ (6B) \\
        & $1.0 \times 10^{-5}$ (12B) \\
        & $6.0 \times 10^{-6}$ (175B) \\
        Learning Rate & $0.1 \times \alpha$ \\
        \midrule
        \textbf{Verification Hyperparameters} & \textbf{Value} \\ 
        \midrule
        Epochs & 2 for generator, 1 for verifier \\
        Sampling Temperature & $0.7$ \\
        Learning Rate & $1.0 \times 10^{-5}$ \\
        Loss weight & $1.0$ \\
        Verifier loss & MSE \\
        Completions per train problem & 100 \\
        Completions per test problem & 100 \\
    \bottomrule
    \end{tabular}
    \caption{Hyperparameters used for all experiments, unless explicitly said otherwise. Notable exceptions include Figure \ref{fig:all_token_dropout}, which uses 4x more tokens per batch and 300 completions at both training and test time. All dropout experiments in Figure \ref{fig:dropout_ablations} use 20\% dropout. Figure \ref{fig:completions} uses verifiers trained on 100 completions, but searching over more completions at test time.}
    \label{table:all_hparams}
\end{table}

\section{Calculator Annotations} \label{appendix:calculator_annotations}

The calculator annotations were not provided by human contractors: they were generated by a combination of hard-coded logic and a finetuned language model. The logic for auto-generating calculator annotations is imperfect. It is highly unlikely to generate any incorrect annotations, but it is not uncommon for it to ignore some lines that could be annotated.

During training, there is no special distinction between the annotated tokens and the rest of the solution: they are all just tokens. During testing, we override model sampling when a well-formatted annotation exists, specifically overwriting the token(s) directly following “=” and within \textless\textless\ldots\textgreater\textgreater.

\begin{figure}[h]
\centering
\begin{subfigure}{0.7 \textwidth}
\includegraphics[width=\textwidth]{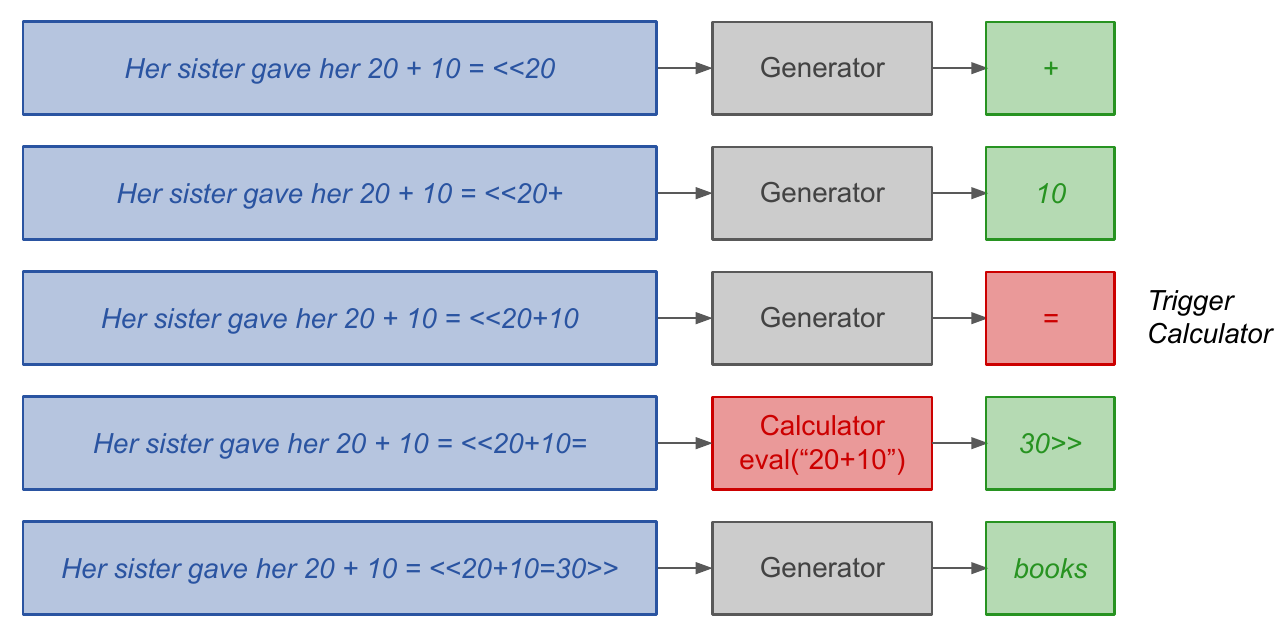}
\end{subfigure}
\caption{A diagram of the calculator sampling procedure.}
\label{fig:calculator_sampling}
\end{figure}

To simulate the calculator, we simply use the python \texttt{eval} function to evaluate the tokens in the expression (Figure \ref{fig:calculator_sampling}). Evaluations that time out or throw an error result in the annotations being skipped and the model being sampled from as usual.

We note that the original version of our calculator, used for all results in this paper, had some minor implementation bugs. Our reported test performance is therefore a slight underestimate, though the magnitude of this discrepancy is less than 1\% in most experiments. Fixing the calculator improves verification test performance by about 1\% when using the full GSM8K training set.

\newpage

\section{Example Model Solutions} \label{appendix:example_solutions}

We showcase a handful of samples comparing finetuning and verification at both 6B and 175B scale. Samples were slightly cherry-picked for diversity.

\begin{figure}[h]
\centering
\begin{subfigure}{.475 \textwidth}
\includegraphics[width=\textwidth]{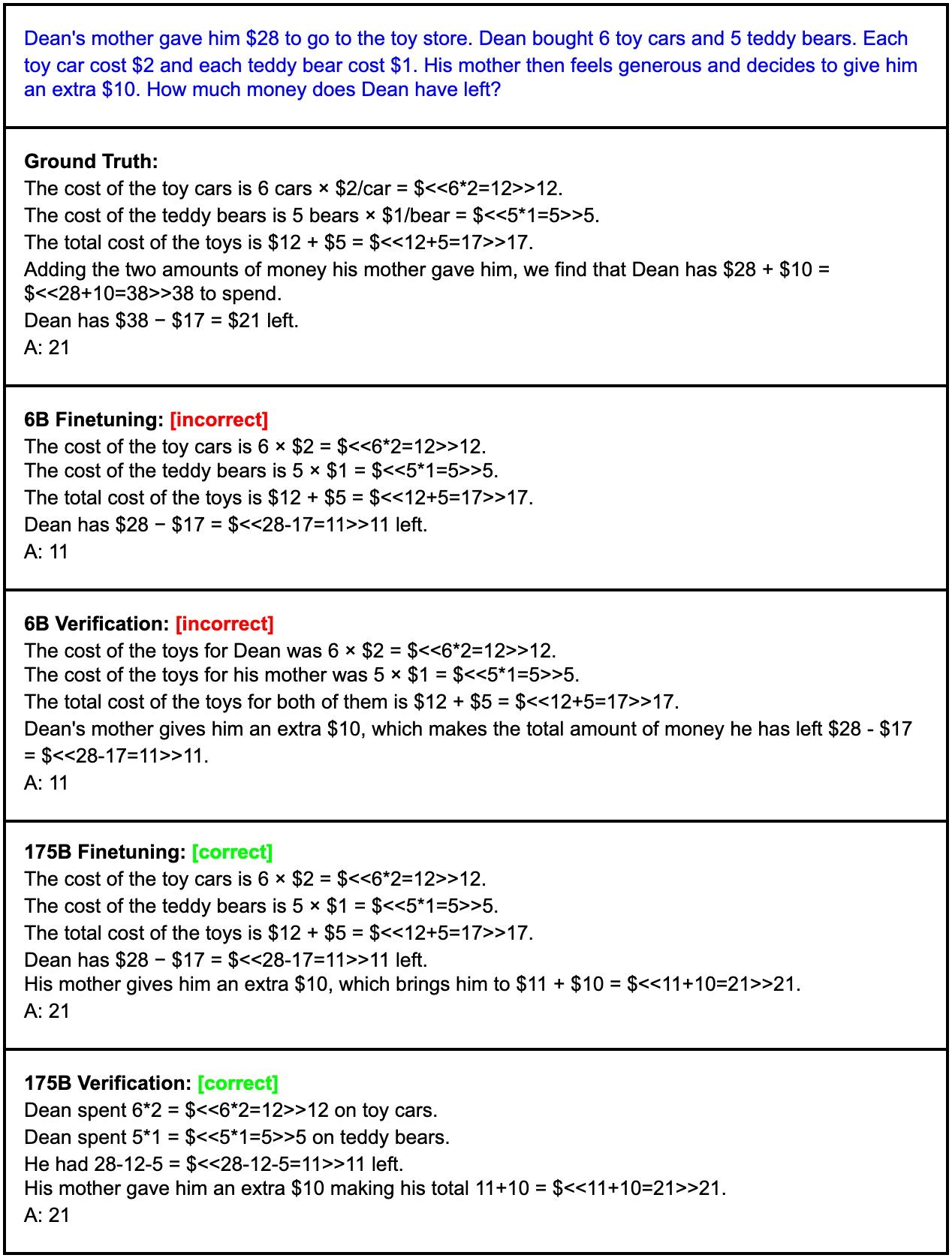}
\vspace*{.75mm}
\includegraphics[width=\textwidth]{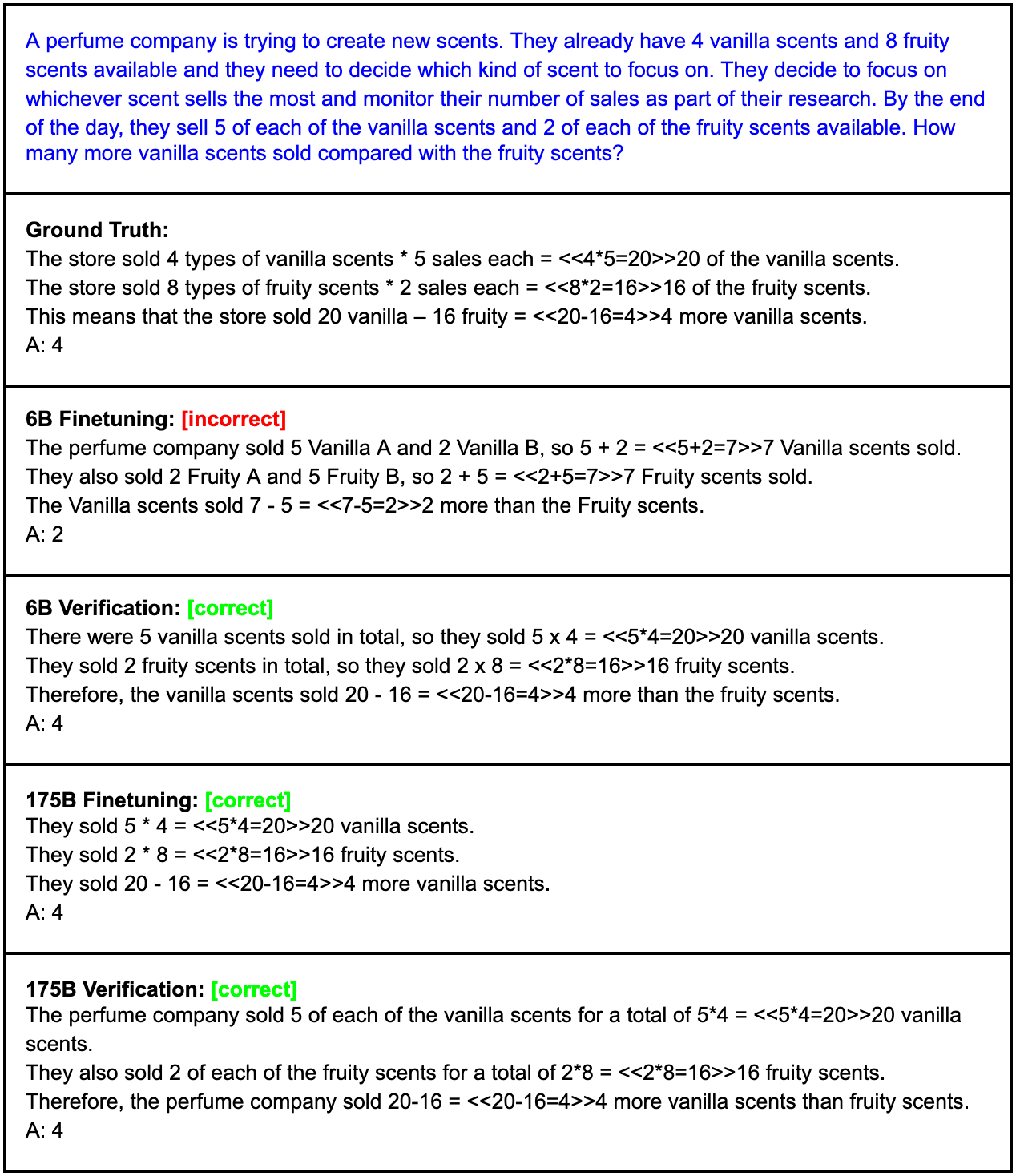}
\end{subfigure}
\begin{subfigure}{.475 \textwidth}
\includegraphics[width=\textwidth]{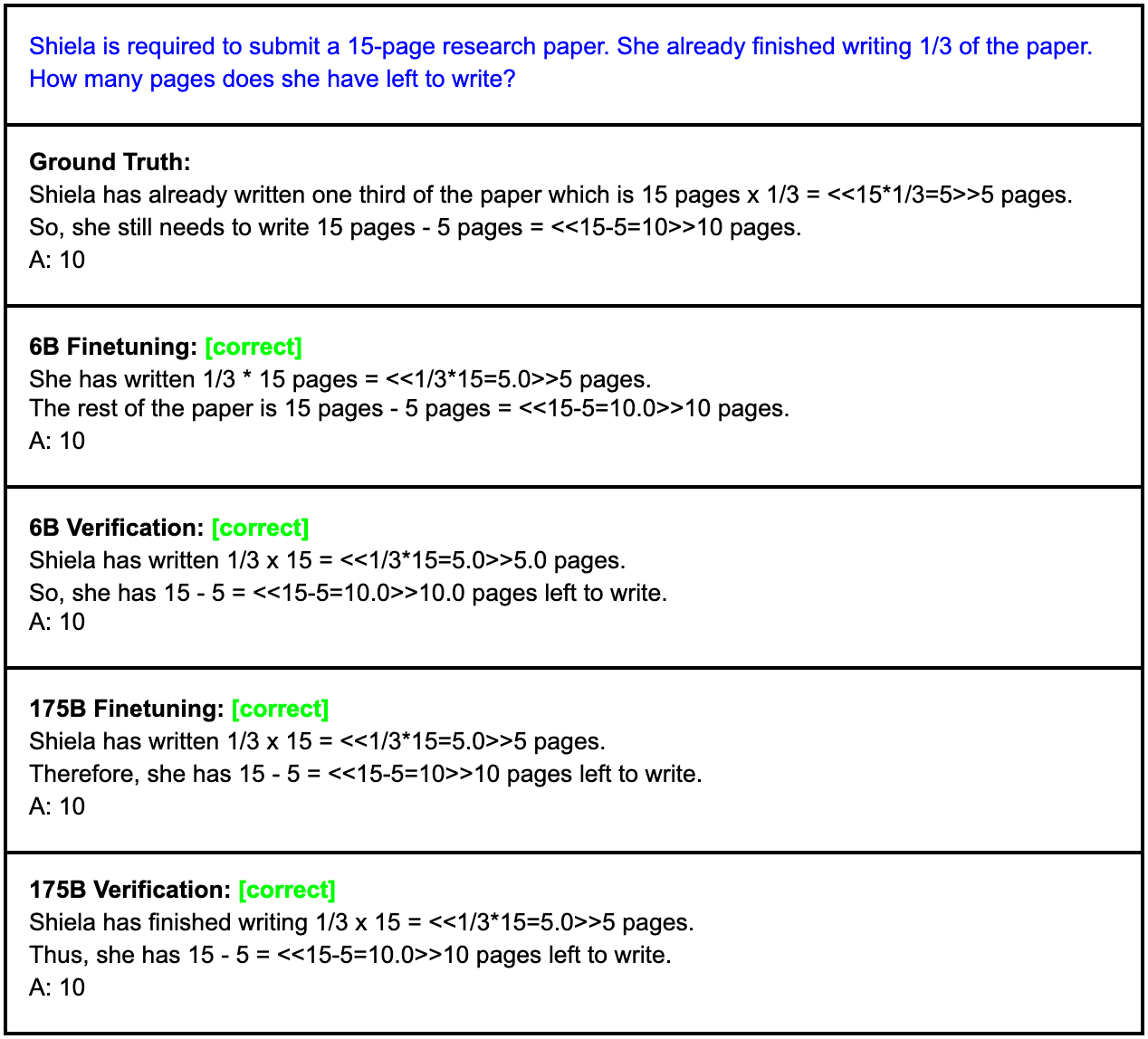}
\vspace*{.75mm}
\includegraphics[width=\textwidth]{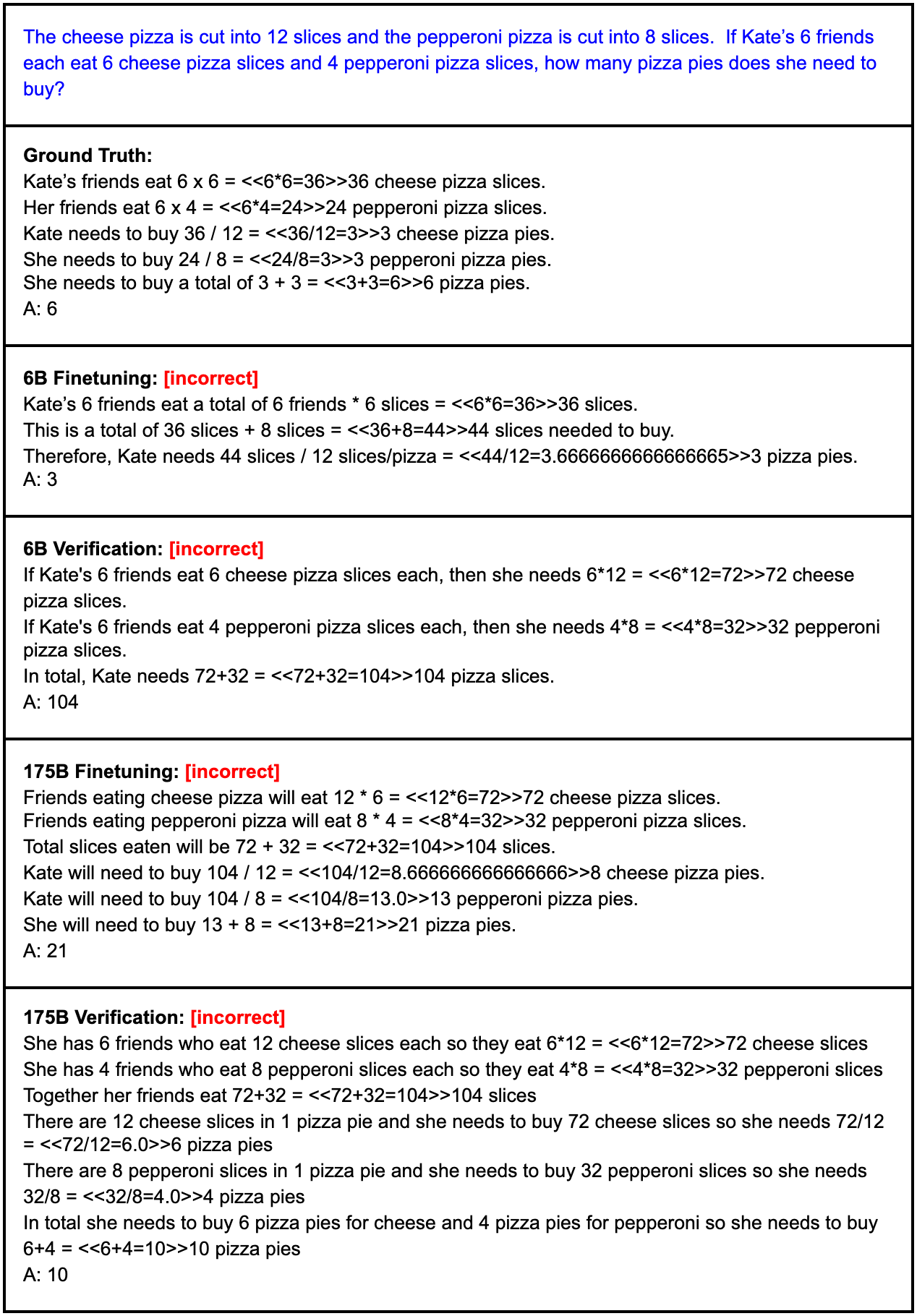}
\vspace*{3.7mm}
\end{subfigure}
\end{figure}

\newpage

\begin{figure}[h]
\centering
\begin{subfigure}{.475 \textwidth}
\includegraphics[width=\textwidth]{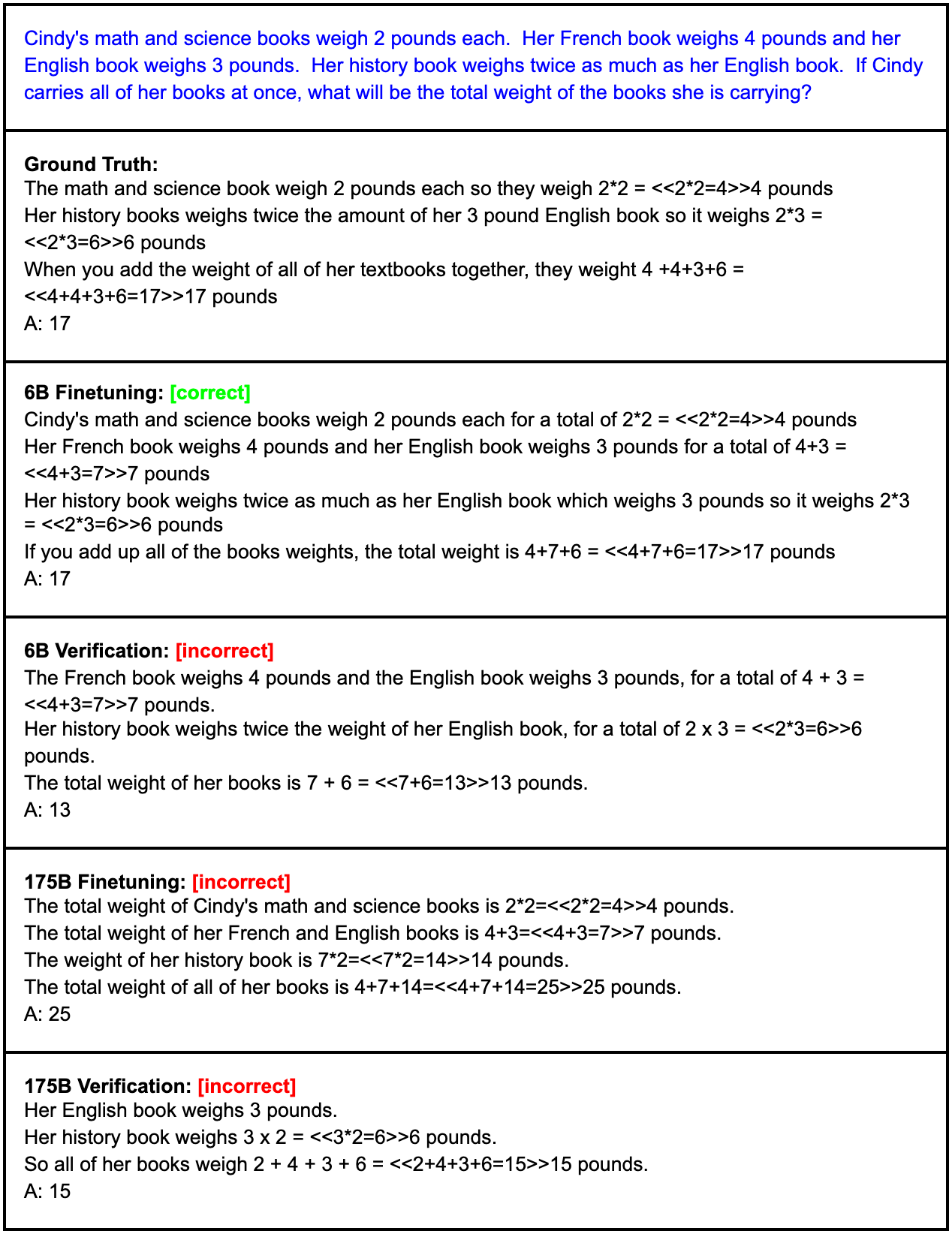}
\vspace*{.75mm}
\includegraphics[width=\textwidth]{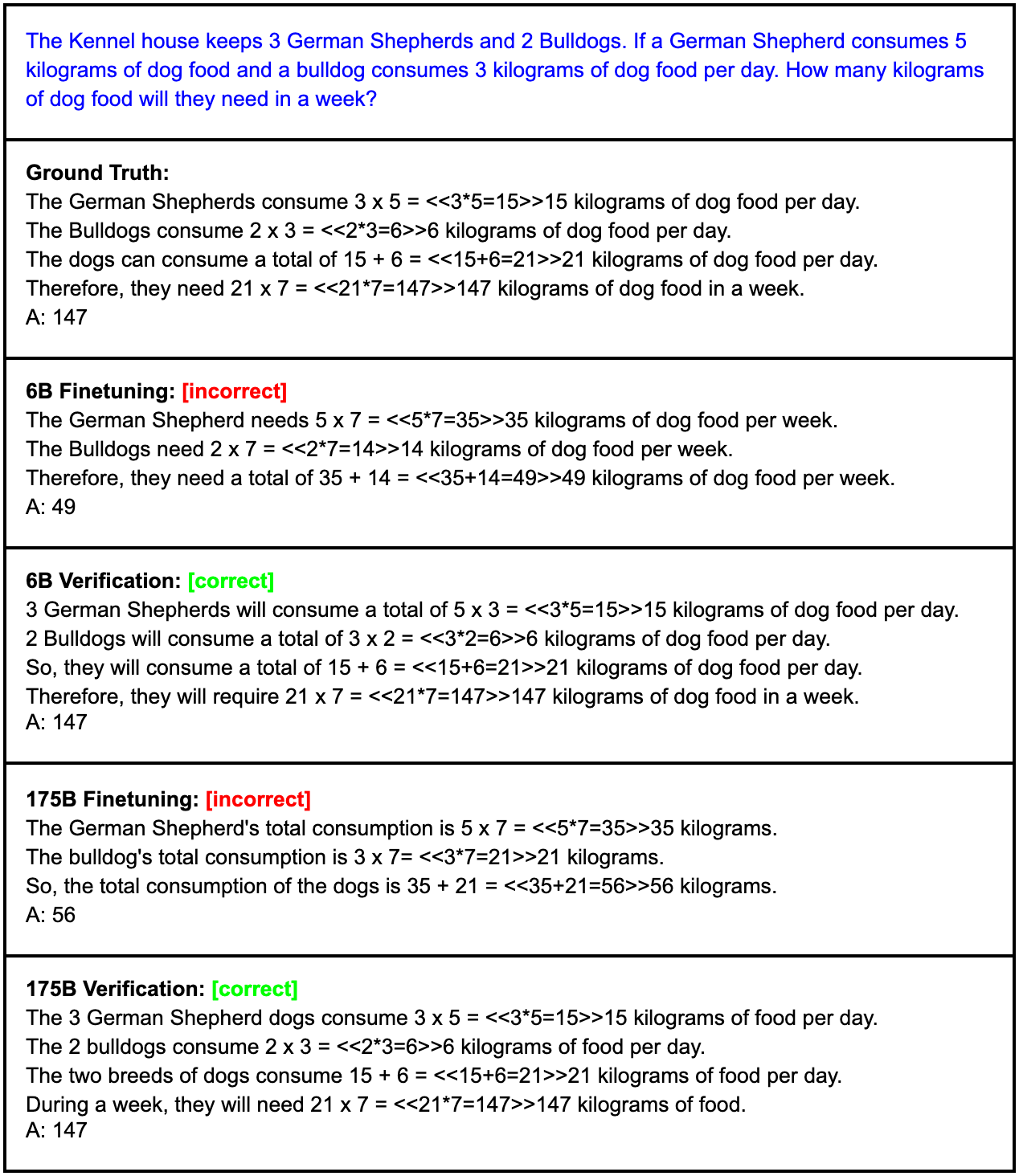}
\end{subfigure}
\begin{subfigure}{.475 \textwidth}
\includegraphics[width=\textwidth]{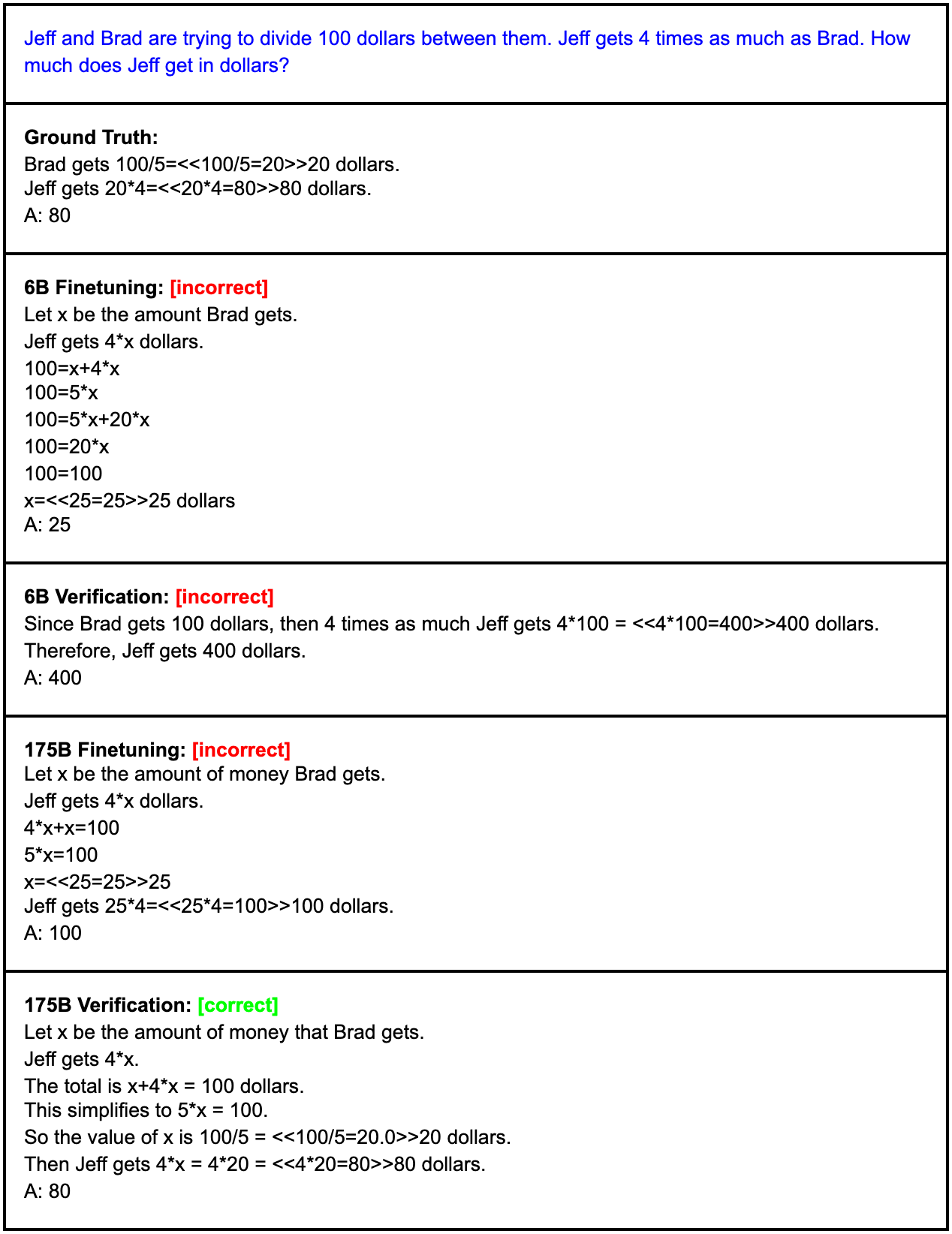}
\vspace*{.75mm}
\includegraphics[width=\textwidth]{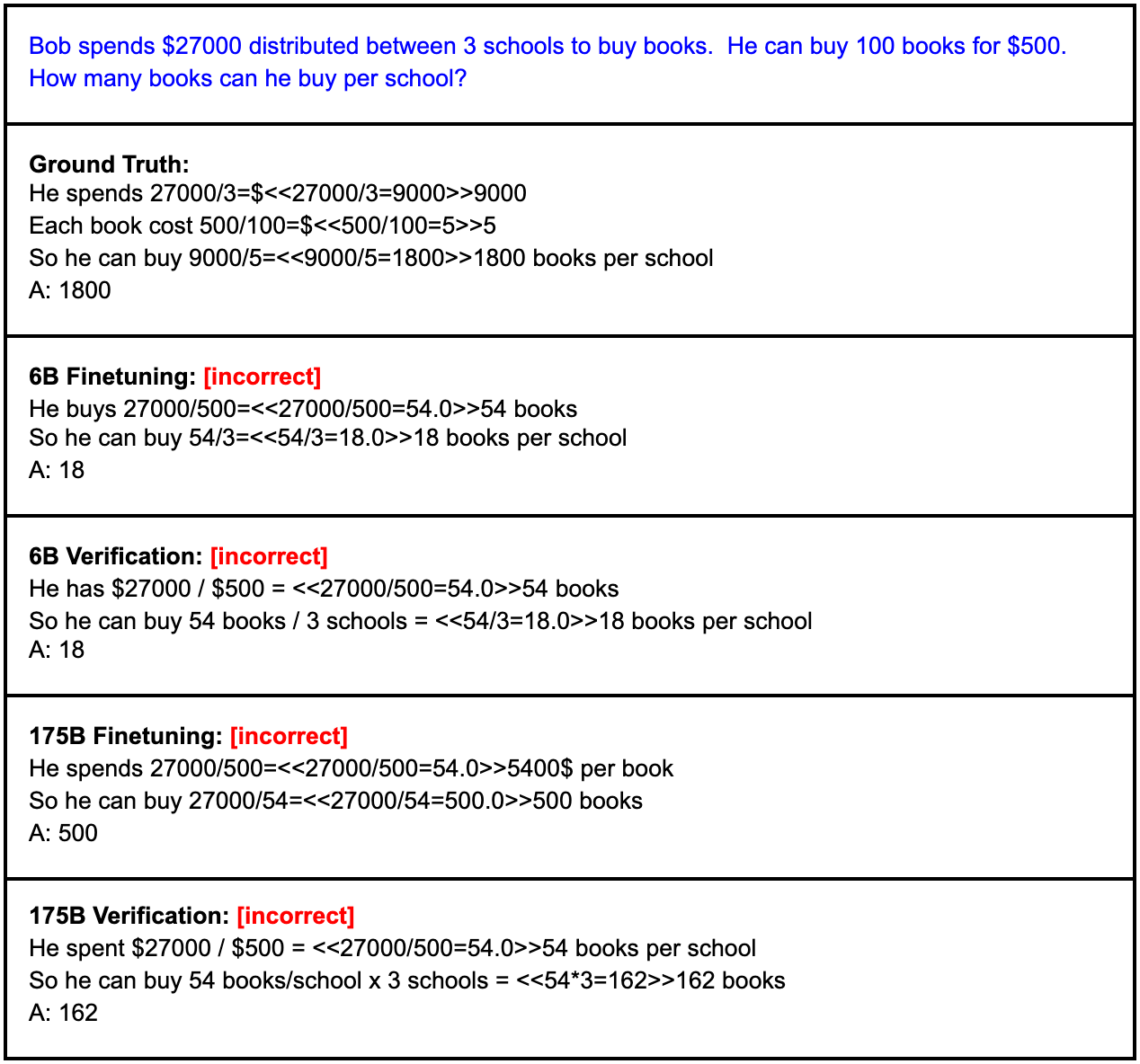}
\vspace*{8.5mm}
\end{subfigure}
\end{figure}

\newpage

\section{Verifier Details} \label{appendix:verifier_details}

As noted in section \ref{section:verification}, we train verifiers with a joint objective where the model learns to label a model completion as correct or incorrect, in addition to the original language modeling objective. Architecturally, this means our verifiers are language models, with a small scalar head that outputs predictions on a per-token basis.

We implement this scalar head as a single bias parameter and single gain parameter that operate on the logits outputted by the language model’s final unembedding layer. Specifically, the bias and gain shift and scale the logit corresponding to a special token in the vocabulary. As such, the logits for other tokens can continue to represent the language modeling objective, while this special token is reserved for the verifier’s predictions.

We can choose to initialize the verifier from the same pretrained language model the generator was finetuned from, or from the generator itself. In our ablations the latter performed slightly better; we suspect this is because better understanding the language distribution that the generator learned should only aid the verifier in scoring samples from that distribution. Unless otherwise explicitly stated, we initialize our verifiers from their corresponding generators in all experiments.

\begin{figure}[h]
\centering
\begin{subfigure}{0.85 \textwidth}
\includegraphics[width=\textwidth]{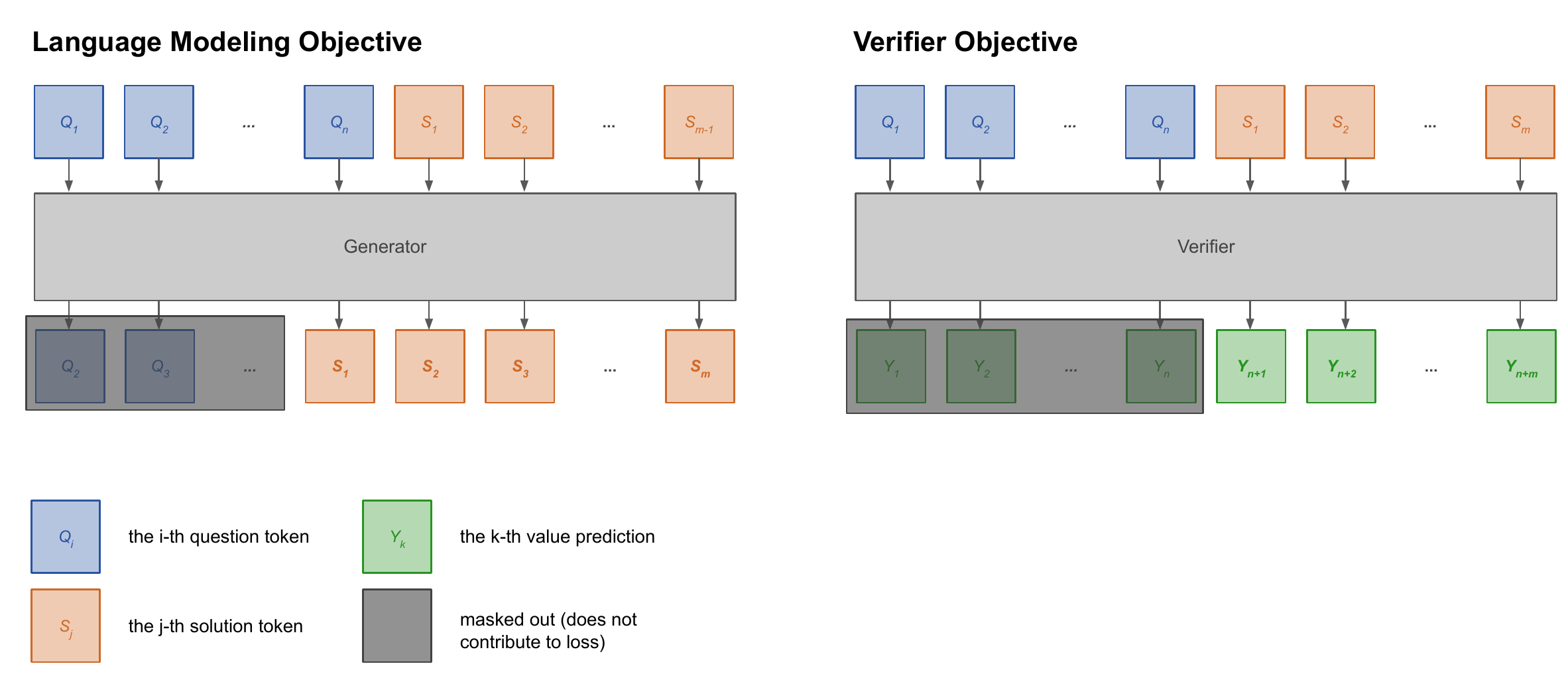}
\end{subfigure}
\caption{Visualization of the joint training objective. We mask out tokens in the question and only consider the loss corresponding to tokens in the solution.}
\label{fig:verifier_masking}
\end{figure}

When training verifiers with the joint objective, we use an equal mix of language data and verifier data. Because we sample 100 completions for each original training example to generate the verifier data, using an equal mix means we effectively upsample the original language data by a factor of 100. To form the joint objective, we simply add the verifier loss and language modeling loss unweighted, and define an epoch of this joint objective as having seen each verifier example once. With both objectives, we mask out tokens in the question and only train on tokens in the solutions, as visualized in Figure \ref{fig:verifier_masking}.

\newpage
\section{Verifier Visualization} \label{appendix:verifier_visualization}

\begin{figure*}[h]
\centering
\includegraphics[width=.75 \textwidth]{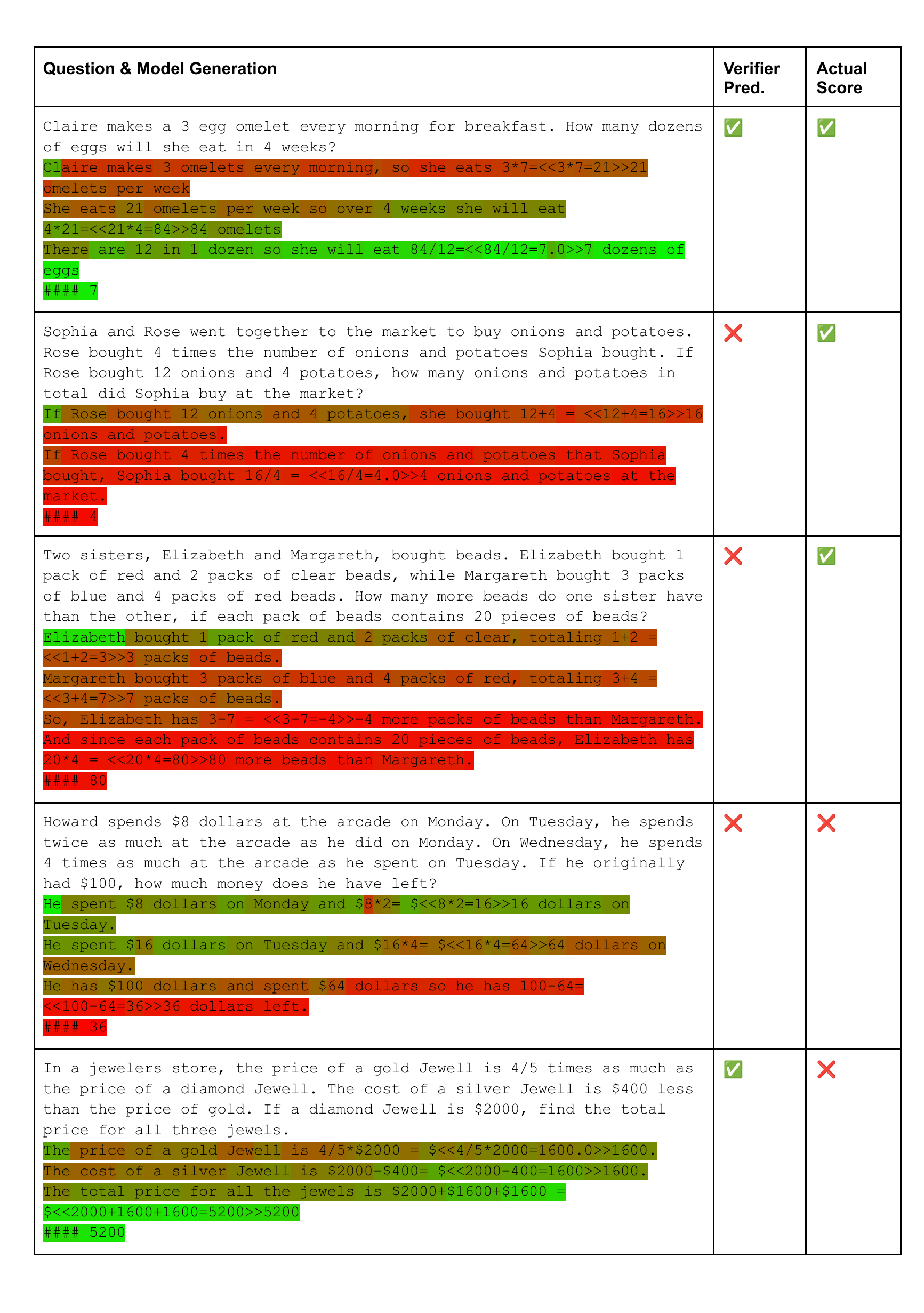}
\caption{Five cherry-picked samples generated by a 175B finetuned model and scored by a 175B token-level verifier. A green background color indicates a high verifier score, and a red background color indicates a low one.}
\end{figure*}

One benefit of the token-level verifiers is that these models become immediately interpretable: we can visualize the predicted value for each token and better understand how the verifier makes decisions on judging samples. Above we present a visualization of the predicted values for five different cherry-picked questions and model completions, verified by a 175B token-level verifier that was trained on the full training set.

In the visualization, the background color of the text corresponds to the verifier score for that token, where red is low value (predicted incorrect) and green is high value (predicted correct). The second column of the table summarizes the verifier's prediction, and the third column indicates whether the generated model completion was actually correct or incorrect. Any disagreement between the second and third columns indicates that the verifier made an error.

The first row includes a true positive example, where the verifier correctly classifies the completion as correct. Note that the model is initially unsure about whether the solution is correct and gradually gains certainty as the solution progresses: this is likely a property of the verifier training procedure, where it trains on a large fraction of incorrect model-generated samples.

The second row contains a problem where the solution is correct, but the verifier has rated it as incorrect. This is potentially due to the ambiguity between the ``4 times'' and the ``4 potatoes'' in the problem description.

The third row consists of another false negative example. However, unlike the previous example, here the model completion contains some faulty reasoning. As such, even though the final answer in the model completion was correct, the natural language explanation was incorrect, and so the verifier correctly assigned a low score.

In the fourth row we see the verifier score a model completion that starts out correct, but where the verifier gradually becomes less confident in the solution as the solution progresses. After the solution makes a clear mistake (saying that \$64 dollars were spent, instead of the $64+16+8=\$88$), the verifier judges the solution as incorrect with a high degree of confidence.

The final row contains a false positive, where the model makes a mistake on the second step, where it subtracts 400 from the price of a diamond jewel instead of a gold one. Verifiers occasionally make mistakes with performing this variable binding of quantities to their relationships.

\end{document}